\documentclass[final,1p,times]{elsarticle}

\usepackage{amssymb}
\usepackage{multirow}
\usepackage{booktabs}
\usepackage{amsmath}
\usepackage[linesnumbered,boxed,ruled,commentsnumbered]{algorithm2e}

\usepackage[table,xcdraw]{xcolor}
\usepackage[colorlinks,linkcolor=red,anchorcolor=blue,citecolor=green]{hyperref}
\usepackage{longtable}
\usepackage{lscape}
\usepackage{graphicx}
\usepackage[table,xcdraw]{xcolor}
\usepackage{subfigure}
\usepackage{amssymb}
\usepackage{multicol}
\usepackage{caption}
\usepackage{changepage}
\usepackage{lipsum}

\makeatletter
\newcommand*{\centerfloat}{%
	\parindent \z@
	\leftskip \z@ \@plus 1fil \@minus \marginparwidth
	\rightskip \leftskip
	\parfillskip \z@skip}
\renewcommand{\@thesubfigure}{\hskip\subfiglabelskip}
\makeatother

\SetCommentSty{mycommfont}

\newcommand{\Autoref}[1]{%
	\begingroup%
	\def\chapterautorefname{Chapter}%
	\def\sectionautorefname{Section}%
	\def\subsectionautorefname{Section}%
	\def\subsubsectionautorefname{Section}%
	\def\algoautorefnam{Algorithm}%
	\def\algorithmautorefnam{Algorithm}%
	\autoref{#1}%
	\endgroup%
}

\newcommand{\ignore}[1]{}

\long\def\comment#1{}

\DeclareCaptionLabelFormat{cont}{#1~#2\alph{ContinuedFloat}}
\begin{document}
	\begin{frontmatter}
		
		

          \title{A Novel Meta-Heuristic Optimization Algorithm Inspired by the Spread of Viruses}
		
		
		\author{Zhixi Li\corref{cor}}
		\ead{lizx@eee.hku.hk}
		\author{Vincent Tam}
		\ead{vtam@eee.hku.hk}
		\cortext[cor]{Corresponding author.}
		\address{Department of Electrical and Electronic Engineering, The University of Hong Kong, Pokfulam, Hong Kong}
		
\begin{abstract}
According to the no-free-lunch theorem, there is no single meta-heuristic algorithm that can optimally solve all optimization problems.
This motivates many researchers to continuously develop new optimization algorithms.
In this paper, a novel nature-inspired meta-heuristic optimization algorithm called virus spread optimization (VSO) is proposed. 
VSO loosely mimics the spread of viruses among hosts, and can be effectively applied to solving many challenging and continuous optimization problems.
In VSO, the ribonucleic acid (RNA) of the virus represents a solution to the problem at hand.
Here, we devise a new representation scheme and viral operations that are radically different
from all the previously proposed virus-based optimization algorithms.
First, the viral RNA of each host in VSO denotes a potential solution for which different viral operations will help to diversify the searching strategies
in order to largely enhance the solution quality.
In addition, an imported infection mechanism, inheriting the searched optima from another colony,
is introduced to possibly avoid the prematuration of any potential solution in solving complex problems.
VSO has an excellent capability to conduct adaptive neighborhood searches
around the discovered local and global optima for achieving better solutions.
Furthermore, with a flexible infection mechanism, VSO is able to quickly
escape from local optima so as to look for other globally (sub-)optimal solution(s).
To clearly demonstrate both its effectiveness and efficiency,
the newly proposed VSO is critically evaluated on a series of well-known benchmark functions.
Moreover, VSO is validated on its applicability through
two real-world examples including the financial portfolio optimization and optimization of hyper-parameters of support vector machines for classification problems.
The experimental results show that VSO has attained superior performance in terms of solution fitness, convergence rate, scalability, reliability, and flexibility when compared to those results of the
conventional as well as state-of-the-art meta-heuristic optimization algorithms.
			
\end{abstract}
		


%
%

		
	\end{frontmatter}
	
	
\section{Introduction}\label{section:introduction}
Optimization techniques have been widely applied in many scientific and engineering applications.
For instance, in the field of artificial intelligence, researchers often attempt to optimize various machine learning models, e.g. tuning hyper-parameters of support vector machines (SVMs)~\cite{RN1} and optimizing deep neural network architecture~\cite{RN2, RN3}, to obtain a better performance. In the areas of industrial design and manufacturing, engineers always encounter numerous optimization problems for various products and scenarios, such as the optimization of aerodynamic shapes for aircraft, cars, bridges, etc.~\cite{RN5} and the optimization of supply chain management~\cite{RN6}.
In finance, investors usually pursue an optimal portfolio aiming to maximize the return while minimizing the risk~\cite{RN10, RN9}. There are many optimization problems in our daily lives like finding the shortest vehicle route to a destination~\cite{RN7}, resource allocation to satisfy performance goals~\cite{RN8}, and so on. 
	
Since many real-world optimization problems are too complex to be solved with a good solution
by conventional optimization approaches in a reasonable time,
meta-heuristic optimization algorithms have recently captured much attention and achieved some success~\cite{RN11}.
In the past decades, researchers have invented several nature-inspired meta-heuristic optimization algorithms
to imitate some phenomena or behaviors of the nature.
Such algorithms can be classified into five categories: evolution-based, swarm-intelligence-based, physics-based, chemistry-based and human-based algorithms.
Evolutionary algorithms (EAs) are inspired by the biological evolutionary process. Genetic algorithm (GA)~\cite{RN12}, evolution strategies (ES)~\cite{RN13} and differential evolution (DE)~\cite{RN14} can be regarded as representative algorithms in EAs.
For the second category, swarm intelligence algorithms (SIs) imitate the intelligent behaviors of creatures in nature. Particle swarm optimization (PSO) is the most pioneering work of SIs~\cite{RN16}. Up to now, the research of SIs has been very active such that new algorithms are being proposed from time to time. Some well-known examples of SIs include: Ant colony optimization (ACO)~\cite{RN15}, artificial bee colony (ABC)~\cite{RN17}, social spider algorithm (SSA)~\cite{RN23}, whale optimization algorithm (WOA)~\cite{RN18}, grey wolf optimizer (GWO)~\cite{RN19}, etc.
For both physics-based and chemistry-based optimization algorithms, that are motivated by physical phenomena and chemical reactions, examples include simulated annealing (SA)~\cite{RN20}, chemical reaction optimization (CRO)~\cite{RN21}, nuclear reaction optimization (NRO)~\cite{RN22} and so on. Lastly, collective decision optimization algorithm (CDOA)~\cite{RN25} and queuing search algorithm (QSA)~\cite{RN24} are examples of the last category.

According to the no-free-lunch theorem (NFL), there is no single meta-heuristic algorithm that can optimally tackle all optimization problems~\cite{RN26}. Undoubtedly, this motivates researchers to continuously develop new algorithms for various applications. In particular, the proposed algorithm should be very competitive with the few existing successful optimization approaches such as PSO for solving the well-known benchmark functions as well as various real-world problems in terms of the solution quality, rate of convergence, scalability, reliability and flexibility, etc.
	
In this paper, we propose a novel, powerful and nature-inspired meta-heuristic algorithm namely the virus spread optimization (VSO) for tackling continuous optimization problems. VSO mimics the mighty spread of viruses among hosts.
Here, we devise a new representation scheme and operations that are radically different from all the previously proposed virus-based optimization algorithms.
First, the viral ribonucleic acid (RNA) of each host in VSO denotes a potential solution to the problem at hand for which different viral infection, mutation and recovery operations will help to diversify the searching strategies
in order to largely enhance the solution quality.
In addition, an imported infection mechanism, inheriting the searched optima from another colony,
is introduced to possibly avoid the prematuration of any potential solution in solving complex problems. 
The VSO algorithm has an excellent capability to conduct adaptive neighborhood searches
around the discovered local and global optima for achieving better solutions.
Furthermore, with a flexible infection mechanism, VSO can quickly escape from local optima in order to look for other globally (sub-)optimal solution(s).

To evaluate the performance of the proposed optimization algorithm, experiments are conducted
on a series of well-known benchmark functions including $16$ classical examples listed in~\cite{RN31}~\cite{RN66}~\cite{RN77} and $30$ problems specially designed by the IEEE CEC 2014 for competition~\cite{RN29}. In addition, VSO is applied to two real-world applications such as the financial portfolio optimization and optimization of hyper-parameters of SVMs for classification problems. To investigate the scalability, the algorithm was well-tested on the classical benchmark functions and portfolio optimization problems with different ranges of dimensions including: low ($30$ \& $100$ dimensions), medium ($300$ \& $500$ dimensions) and high ($1,000$ dimensions) for the benchmark functions, and different numbers as $30$, $100$ and $250$ of stocks for portfolio optimization. A standardized running environment and settings are used for a fair comparison of the performance of the VSO algorithm with those of the conventional meta-heuristic algorithms including GA~\cite{RN12}, DE~\cite{RN14}, PSO~\cite{RN16}, ABC~\cite{RN17},  as well as state-of-the-art ones, i.e. SSA~\cite{RN23}, WOA~\cite{RN18} and covariance matrix adaptation evolution strategy (CMA-ES)~\cite{RN32} with their outstanding performance reported in literature.
The experimental results verify that VSO achieves impressive performances
in terms of solution quality, convergence rate, scalability, reliability and flexibility when compared to those of
the above conventional and state-of-the-art meta-heuristic algorithms.
	
To the best of our knowledge, two virus-based algorithms namely the virus colony search (VCS)~\cite{RN28} and virus optimization algorithm (VOA)~\cite{RN27} have been proposed to tackle various optimization problems.
However, VSO is radically different from these two existing algorithms in their analogies, motivation, implementations and search behaviors. We will further reveal the details in~\Autoref{section:vso}.
	
In summary, the major contributions of this paper are as follows.	
\begin{itemize}
	\item  A new meta-heuristic algorithm as a very competitive and potential approach is proposed to solve challenging and continuous optimization problems;
	\item The proposed optimization algorithm, combining the strengths of EAs and SIs, can achieve an excellent trade-off between exploitation and exploration by the unique design of the diversification of the search strategies. This makes the algorithm applicable to a wider range of problems in practice;
	\item  The imported infection mechanism, as a novel search strategy cooperating with other meta-heuristic algorithms, helps to significantly enhance the overall optimization algorithm for tackling more complex problems;
	\item The outstanding performance of the proposed algorithm is demonstrated not only on the solution quality but also the rate of convergence, scalability and reliability through performing a series of experiments on $46$ well-recognized benchmark functions and two real-world optimization problems.
\end{itemize}
	
The rest of this paper is organized as follows.
~\Autoref{section:vso} describes the analogies, operations, implementations and work flow of the VSO algorithm in details.
The experimental results and related discussion on the benchmark functions are presented in~\Autoref{section:benchmark_function_evaluations}.
The performances for two real-world applications including financial portfolio optimization and optimization of hyper-parameters of SVMs for classification problems are shown and discussed in \Autoref{section:portfolio_optimization} and~\Autoref{section:SVM} respectively. We conclude this work and shed lights on various potential future directions in~\Autoref{section:conclusion}.
	
\section{The Virus Spread Optimization Algorithm}\label{section:vso}
\subsection{Analogies and Definitions}
Considering the powerful spread of viruses with a great diversity of viral behaviors,  VSO is proposed to simulate such process loosely. The analogies of VSO are listed in ~\autoref{tab:analogy}. The host and virus are essential components of the algorithm. 
	
\begin{table}[h]
    \footnotesize
    \centerfloat
	\caption{Analogy of VSO}
	\label{tab:analogy}
	\begin{tabular}{@{}lll@{}}
		\toprule
		Terminology &
		\multicolumn{1}{c}{Natural Meaning} &
		\multicolumn{1}{c}{Algorithmic Meaning} \\ \midrule
		Viral Spread &
		To infect all hosts. &
		\begin{tabular}[c]{@{}l@{}}To search the solution space and find an optimal one.\end{tabular} \\ \midrule
		Virus &
		\begin{tabular}[c]{@{}l@{}}A virus that contains an RNA which may mutate.\end{tabular} &
		\begin{tabular}[c]{@{}l@{}}The RNA represents a solution to the problem.\end{tabular} \\ \midrule
		Host &
		\begin{tabular}[c]{@{}l@{}}Organism (e.g. animals, humans) that is infected \\ by the virus. The infected host may show symptoms \\ of various degrees. \end{tabular} &
		\begin{tabular}[c]{@{}l@{}}The symptom intensity generally represents the \\ fitness of a feasible solution. The critical host \\ denotes the best   fitness.\end{tabular} \\ \bottomrule
	\end{tabular}
\end{table}

In VSO, the population is composed of hosts. There are four types of hosts imitating the spread of viruses and the immunological differences in nature: {\em healthy, mild, severe} and {\em critical\/}. Each host including the healthy one carries a virus. 
In fact, many animals including humans may carry all kinds of non-infectious viruses in nature~\cite{RN35}. For instance, a healthy human may carry a few viruses like endogenous retrovirus (ERV) that are in fact beneficial to our immune system~\cite{RN34}. Besides, bats carry a lot of unknown viruses yet may not get sick from those viruses~\cite{RN33}. 
The main difference between $healthy$ hosts and other hosts is that $healthy$ hosts act as healthy carriers with non-infectious viruses while the infected (also called infectious) ones, i.e. {\em mild, severe, and critical\/} hosts, can infect the $healthy$ hosts. 

More importantly, there are different viral infection and mutation operations for each type of hosts in VSO to diversify the searching strategies so that the optimizing capability and flexibility can be largely enhanced. More definitions are provided as follows.

\begin{itemize}
	\item Definition 1: A Viral RNA\\
	Each host has a viral RNA that represents a possible solution as shown in~\eqref{eq:eq1}.
	\begin{equation}\label{eq:eq1}
	X_i=[x_i^1,x_i^2,\ldots,x_i^D]
	\end{equation}
    where $ X_i$ (vector) is the RNA of the virus denoting a possible solution to the problem at hand, $i$ is the iteration number, and $D$ is the dimensionality, i.e. the number of decision variables, of the problem.
			
	\item Definition 2: A Healthy Host\\		
	A $healthy$ host is a host carrying a non-infectious virus whose RNA is generated randomly in every iteration. The host conducts a random search in the solution domain as listed in~\eqref{eq:eq2}.
	\begin{equation}\label{eq:eq2}
	X_i=\ U(S)
	\end{equation}
    where $S$ is the whole search space  while $U$ is a random number generator function based on the uniform distribution of $S$.
  
	\item Definition 3: A Mild Host\\	
	A $mild$ host is carrying an infectious virus. As shown in~\eqref{eq:eq3}-\eqref{eq:eq4}, the virus of this host can mutate with a mutation intensity $intensity_i^M$ and also infect other $healthy$ hosts with a rate $R^M$ that is relatively low when compared to other infectious hosts. 
	\begin{equation}\label{eq:eq3} {intensity}_i^M=\alpha\ast{intensity}_{i-1}^M+\gamma\ast\ rand\left(0,1\right)\ast\ ({gbest}_{i-1}-X_i)                          
	\end{equation}
	\begin{equation}\label{eq:eq4}
	X_{i+1}=\ X_i+\ {intensity}_i^M
	\end{equation}	
	where $intensity_i^M$ (vector) is the mutation intensity of the $mild$ host at the iteration $i$, $\alpha\in [0,1]$ and $\gamma\in[1,2]$ are the scaling factors, ${gbest}_{i-1}$ is the best solution obtained by the population at the iteration $i-1$, and $rand(0,1)$  is a random number between $0$ and $1$.
		
	\item Definition 4: A Severe Host\\	
	As shown in~\eqref{eq:eq5}-\eqref{eq:eq6}, a $severe$ host carries an infectious virus that can mutate with a mutation intensity $intensity_i^S$ and also infect other $healthy$ hosts with its own rate $R^S$. Overall speaking, its infectious ability is medium as compared to that of the $critical$ host.
	\begin{equation}\label{eq:eq5}
	{intensity}_i^S=\delta_s\ast{intensity}_{i-1}^S\\
	\end{equation}
	\begin{equation}\label{eq:eq6}
	X_{i+1}=\ X_i+\ Gaussian(0,{intensity}_i^S) * \ X_i\
	\end{equation}
    where ${intensity}_i^S$ (scalar) is the mutation intensity  of the $severe$ host at the iteration $i$, $\delta_s \in (0,1]$ is the decay rate, and $Gaussian(0,{intensity}_i^S)$ is the Gaussian function with the mean as $0$ and the standard deviation as ${intensity}_i^S$.

	\item Definition 5: A Critical Host\\	
	In VSO, there is only one $critical$ host which represents the currently most optimal solution obtained so far. As shown in~\eqref{eq:eq7}, its viral mutation is paused yet with the highest infection rate $R^C$ to carry its relatively good solution quality to other $healthy$ hosts.
	\begin{equation}\label{eq:eq7}
	X_{i+1}=\ X_i 
	\end{equation}
\end{itemize}
	
\subsection{Operations}\label{section:operations}
In VSO, the {\em initialization, selection, mutation, infection\/} and {\em recovery\/} are five essential operations while the  {\em imported infection\/} serves as an additional operation to enhance the optimizing performance.

\subsubsection{Initialization}\label{section:initialization}
At the starting point with the number of iterations as $0$, the whole population is initialized as $healthy$ hosts. The viral RNA of each host is randomly generated in the search space according to~\eqref{eq:eq8}.	
\begin{equation}\label{eq:eq8}
X_{i=0}=bound_{l}  + rand(0,1) * (bound_{u} - bound_{l})  
\end{equation}
where $i$ is the iteration number, $bound_{l}$ and $bound_{u}$ denote the lower and upper bounds of the corresponding domain of the variable being considered.

The mutation intensities  ${intensity}_i^S$ and ${ intensity}_i^M$ of the mild and $severe$ hosts are initialized in~\eqref{eq:eq9} and~\eqref{eq:eq10} as below.
\begin{equation}\label{eq:eq9}
intensity^M_{i=0} = \frac{U(bound_{l},bound_{u})}{10}
\end{equation}
\begin{equation}\label{eq:eq10}
intensity^S_{i=0} = \frac{1}{rand(0,1)}
\end{equation}

Algorithm~\ref{algo:initialization} shows the detailed initialization process.

\begin{algorithm}[!htb]
	\footnotesize
	\caption{Initialization}
	\label{algo:initialization} 
	\LinesNumbered
	\KwIn{Population size: $N_{pop}$\newline
	Searching bound: $bound_{l}$ and $bound_{u}$\newline
	Random number generator based on the uniform distritbution: $U$
	}
    \KwOut{Newly created hosts: $hosts$}
    $hosts \gets \emptyset$\;
    $i \gets 0$\;
    \While{$(i < N_{pop})$}{
	    Initialize a new host $h$ with a viral RNA according to \eqref{eq:eq8}\;
	    Initialize the mutation intensities of the host $h$ according to \eqref{eq:eq9} \& \eqref{eq:eq10}\;
    	$h$.type $\gets$ 'healthy'\;
	    Insert $h$ into $hosts$\;
	    $i \gets i + 1$\;
    }

    \Return hosts
\end{algorithm}

\subsubsection{Selection}\label{section:selection}
In VSO, the host with the best solution will be selected as the $critical$ host after calculating fitness for all hosts at each iteration. As presented in Algorithm \ref{algo:selection}, the host that has achieved the best solution up to current iteration will be designated as the $critical$ host while the previous $critical$ one will be downgraded to the $severe$ host. 

In nature, due to the complicated viral mutation, immune response and outside environment, some viruses infecting a $healthy$ host may develop into deadly viruses shortly. Analogously, a $healthy$ host conducting a random search will possibly become the $critical$ one directly as well in VSO.

\begin{algorithm}[!h]
	\footnotesize
	\caption{Selection}
	\label{algo:selection} 
	\LinesNumbered
	\KwIn{Hosts: $hosts$ \newline
		  Current number of iterations: $i$
    }
	\KwOut{critialHost}
    
    Get the host $gBestHost$ with best solution from hosts\;
    $gBestHost.type \gets$ 'critical'\;
    \If{$(gBestHost \neq prev\_gBestHost)$}{
        prev\_gBestHost.type $\gets$ 'severe'\;
        $gBest_i$ $\gets$ gBestHost.virus.rna\;
    }

\end{algorithm}

\subsubsection{Mutation}\label{section:mutation}
The mutation behavior of the searching strategy is one of the key factors to the success of VSO. Depending on the type of hosts, the mutation operation will work according to~\eqref{eq:eq2}-\eqref{eq:eq7}. Algorithm \ref{algo:mutation} clearly shows the pseudo-code of the mutation operation. The viral RNAs of all hosts will be updated accordingly by the mutation operation at each iteration.

\begin{algorithm}[h]
	\footnotesize
	\caption{Mutation}
	\label{algo:mutation} 
	\LinesNumbered
	\KwIn{All hosts: $hosts$}
	\KwOut{All hosts with updated viral RNAs: $hosts'$}
	\For{each host in $hosts$}{
		\Switch{host.type}{
			\Case{healthy}{
				Update host.virus.rna according to \eqref{eq:eq2}\;
			}
			\Case{mild}{
				Update host.virus.$intensity^M$ according  to \eqref{eq:eq3}\;
				Update host.virus.rna according to \eqref{eq:eq4}\;
			}
			\Case{severe}{
				Update host.virus.$intensity^S$ according  to \eqref{eq:eq5}\;
				Update host.virus.rna according to \eqref{eq:eq6}\;
			}
			\Case{critical}{Update host.virus.rna according to \eqref{eq:eq7}\;}
		}
		
	}
\end{algorithm}

\subsubsection{Infection}\label{section:infection}
The main objective of the infection mechanism is to spread the viral information among all the hosts so as to empower the search effectiveness of the VSO algorithm. In the real world, the transmission route, such as direct contact, is necessary for the spread of many viral diseases~\cite{killingley2013routes}. We hereby design a three-step mechanism for the infection operation in VSO.
 
At first, every infectious host has one or more chances to contact $healthy$ hosts at each iteration.

Secondly, we have to decide whether that contacted $healthy$ host will be infected or not. Therefore, different infection rates are assigned to the hosts according to their types as shown in~\eqref{eq:eq11}.

\begin{equation}\label{eq:eq11}
R_{infect} = [R^M,R^S,R^C] 
\end{equation}
where $ 0<R^M \leq R^S < R^C <1$. They are the infection rates for $mild$, $severe$, and $critical$ host, respectively. More specifically, the infection rate is the probability of an infectious host infecting a $healthy$ host when they contact.

Lastly, in case of a $healthy$ host infected by an infectious host successfully, it will become a $severe$ or $mild$ host at different probabilities. We hereby design a transformation matrix as illustrated in~\eqref{eq:eq12}.
\begin{equation}\label{eq:eq12}
P_{trans} = 
\left[
\begin{matrix} 
P^C_{H->M} & P^C_{H->S}  \\ \\
P^S_{H->M} & P^S_{H->S}  \\ \\
P^M_{H->M} & P^M_{H->S} 
\end{matrix}
\right]
\end{equation}
where $P_{trans}$ is the matrix of transformation probabilities. For instance, $P^S_{H->M}$ is the conditional probability of a $healthy$ host becoming the $mild$ host 
given by being infected by a $severe$ host. As there are only two events here that are mutually exclusive, i.e. becoming a mild or severe host, the summation of each row of probabilities is equal to $1$. 

\vbox{}
At each iteration, the specific procedure of the infection is summarized as follows. 

\begin{figure}[h]
	\centering
	\includegraphics[scale=0.5]{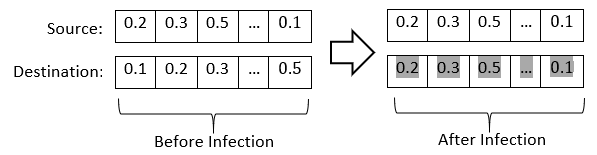}
	\caption{The host to be infected as the $severe$ host}
	\label{fig:infected_as_severe_host}
\end{figure}

\begin{figure}[h]
	\centering
	\includegraphics[scale=0.5]{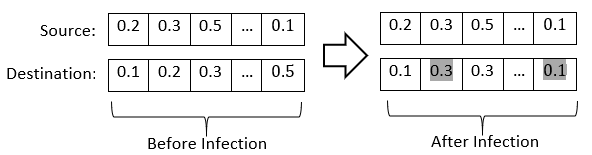}
	\caption{The host to be infected as the $mild$ host}
	\label{fig:infected_as_mild_host}
\end{figure}

\begin{itemize}
	\item As indicated in~\eqref{eq:eq11}, the $healthy$ host contacting with an infectious host will be infected with probabilities  $R^C$, $R^S$and $R^M$ respectively as dependent on the type of the infectious host;
	\item During the infection, the $healthy$ host may be infected as the severe or $mild$ host according to the transformation probabilities as described in~\eqref{eq:eq12}. Specially, the host infected by the $mild$ host can become the $mild$ host only so that the transformation probability $P^M_{H->M}$ is always 1 and $P^M_{H->S}$ is equal to $0$;	
	\item In addition, two solution sharing mechanisms may be performed during the infection process. When a $healthy$ host (destination) infected by an infectious host (source) to become a $severe$ host, the viral RNA of the source will be copied to the destination directly as shown in ~\autoref{fig:infected_as_severe_host}. In addition, when a $healthy$ host is infected by a $mild$ host, each assigned value of the viral RNA of the destination will be randomly replaced by the source with a fixed probability of $0.5$ as shown in ~\autoref{fig:infected_as_mild_host}.

\end{itemize}

The implementation of the above viral infection is described in Algorithm~\ref{algo:infection}. The infectious and $healthy$ hosts are firstly sorted according to the ascending and descending order of their fitness values respectively. Since the VSO algorithm is designated for solving minimization problems, the smaller the fitness value, the better the solution. Thus, the infectious host with a better solution quality will be more likely selected to infect a $healthy$ host. Conversely, a $healthy$ host with a worse fitness value will be more likely to be infected. Moreover, an integer parameter $H$ as mentioned above is used to limit the maximum number of $healthy$ hosts to be contacted by each infectious host. This can help to avoid any premature convergence of the whole population to any local minimum.

\begin{algorithm}[!h]
	\footnotesize
	\caption{Infection}
	\label{algo:infection} 
	\LinesNumbered
	\KwIn{All hosts: $hosts$\newline 
		The maximum number of healthy hosts to be contacted by each infectious host: $H$\newline
		Problem dimensionality: $D$
	}
	\KwOut{All hosts with possibly updated viral RNAs: $hosts'$}
	Select $infectiousHosts$ from $hosts$\;
	Sort $infectiousHosts$ by the ascending order of fitness values\;
	\For{each $infectiousHost \in infectiousHosts$}{
		Select $healthyHosts$ from $hosts$\; 
		\If{$(| healthyHosts | \geq H)$}{
			Sort $healthyHosts$ by the descending order of fitness values\;
			$contactedHosts \gets \{ \mbox{the first $H$ hosts} \} \subseteq healthyHosts$\;
			
				\For{each $healthyHost \in contactedHosts$}{
					$infected \gets false$\;
					$T \gets infectiousHost.type$\;
					Get the infection rate $R^T$ from \eqref{eq:eq11}\;
					\If{rand(0,1) $\leq$ $R^T$}{
						$infected \gets true$\;
			        }
					\If{$(infected)$}{
						Get the transformation probabilities $\langle P^T_{H->M}, P^T_{H->S} \rangle$ from \eqref{eq:eq12}\;
				       \eIf{$(0\leq rand(0,1) \leq  P^T_{H->M})$}{
					      $to\_be\_infected\_type \gets M$\;
				        }{
				          \tcp{$P^T_{H->M}$ + $P^T_{H->S}$ = 1}
					      $to\_be\_infected\_type \gets S$\;
				        }
			            \Switch{$to\_be\_infected\_type$}{
			        	    \Case{$M$}{
			        		    $healthyHost.type \gets$ 'mild'\;
			        		    \For{$idx=0$ to $D-1$}{
			        			    \If{$rand(0,1) \leq 0.5$}{
			        				    $healthyHost$.virus.rna[$idx$] $\gets$ $infectiousHost$.virus.rna[$idx$]\;
			        			    }
			        		    }
			        	    }
			        	    \Case{$S$}{
			        		    $healthyHost$.type $\gets$ 'severe'\;
			        		    $healthyHost$.virus.rna $\gets$ $infectiousHost$.virus.rna\;	
			        	    }  
			            }  
			        }
				}

		}
	}
\end{algorithm}

\subsubsection{Recovery}\label{section:recovery}
	
The recovery operation is another key mechanism of the VSO algorithm. Due to the powerful viral spread in the infection, all hosts may be infected very soon so that the searching capacity of the algorithm may still be quickly converged into a local minima even though with the aforementioned parameter $H$ to restrain the maximum number of contacted hosts. 
	
Thus, in case all the hosts are infected, the recovery operation will be performed to carefully reset some of the infected hosts to continue with the exploration process. We have not adopted the simple random or scheduled restart approaches used by many algorithms such as~\cite{RN44,RN45,RN46}. Instead, an interesting mechanism to gradually downgrade the infected hosts is devised as inspired by the nature in which an infected host has to recover gradually. Likewise, each infected host will be downgraded to the less severe host type of the VSO framework. For example, a $severe$ host will be recovered to the $mild$ host while a $mild$ host will become the $healthy$ host. As the searching restrictions will be relaxed for the ``recovered'' host types, the searching capacity of the algorithm will be enhanced gradually as well so as to explore the other parts of the search space.

Furthermore, a parameter $recPercent$ called the recovery rate is used to specify the percentage of the infected hosts with the worst solution quality to be recovered. This can help to avoid losing all the search information accumulated so far during the search process. The detailed implementation is given in Algorithm~\ref{algo:recovery}.
	
\begin{algorithm}[h]
	\footnotesize
	\caption{Recovery}
	\label{algo:recovery} 
	\LinesNumbered
	\KwIn{Infectious hosts: $infectiousHosts$\newline
		Population size: $N_{pop}$\newline
		Recovery percentage: $revPercent$ 
	}
	\KwOut{Hosts recovered from $infectiousHosts$}
	Sort $infectiousHosts$ by the descending order of fitness values\;
	\If{$(|infectiousHosts| = N_{pop})$}{
		$revNum \gets N_{pop} * revPercent$ \;
		$RH \gets \{ \mbox{the first $revNum$ hosts} \} \subseteq infectiousHosts$\;
		\For{each $host \: h \: \in \:  RH $}{
			Initialize $h$ referring to Algorithm \ref{algo:initialization} \;
			\Switch{$h$.type}{
				\Case{severe}{
					$h$.type $\gets$ 'mild' \;
				}
				\Case{mild}{
					$h$.type $\gets$ 'healthy' \;
				}
			}
		}
	}
\end{algorithm}
	
\subsubsection{Imported Infection}\label{section:imported_infection}
As inspired by the possible migration of hosts from one place to another that may increase the spread of a viral disease in the real world, the concept of ``imported infection'' is introduced as an additional operation of the VSO framework to enhance its search performance for solving complex optimization problems.

Accordingly, a new colony is developed through the DE algorithm to construct some potentially better solution to the whole population of VSO. However, this simple heuristic operation may break the searching patterns of the concerned VSO algorithm, thus possibly leading to a poorer performance. Therefore, an adaptive probability is predefined to export the DE colony to the whole population of VSO in a probabilistic manner as illustrated in Algorithm \ref{algo:imported_infection}.

\begin{algorithm}[!htb]
	\footnotesize
	\caption{Imported Infection}
	\label{algo:imported_infection} 
	\LinesNumbered
	\KwIn{Critial host: $criticalHost$\newline
		DE algorithm: $DE$ with the populaztion size as $N_{im}$ (refer to \cite{RN50}) \newline
		Infection probability: $P_{im}$ \newline
		Current number of iterations: $i$\newline
		Total number of iterations: $j$
	}
	\KwOut{A critical host with the updated viral RNA: $criticalHost$}
	$\langle bestSolution, bestFitness \rangle \gets DE$ \;
	\If{$(rand(0,1) \leq (P_{im} * i / j))$}{
		\If{$(bestFitness < criticalHost.fitness)$}{
			$criticalHost$.virus.rna $\gets bestSolution$
		}
	}
\end{algorithm}

As an additional operation, the imported infection may help to improve the search performance in some complex cases yet it will also increase the overall computational complexity of the VSO algorithm. Hence, we may flexibly skip this additional operation in some cases. More importantly, this novel design provides a useful interface for researchers or users to integrate their own algorithms for some specific problems.

\subsection{The Algorithmic Flow of VSO}\label{section:algorithmic_flow}
\begin{figure}[!htb]
\centerfloat
\includegraphics[scale=0.7]{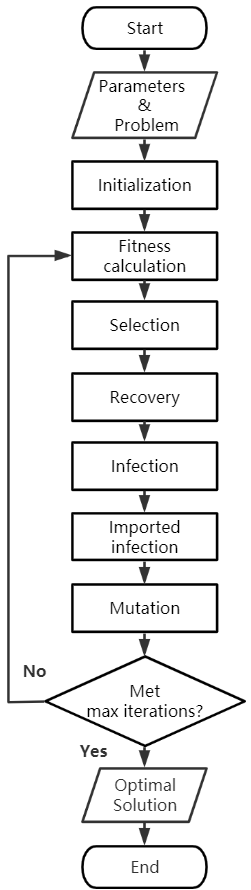}
\caption{The Algorithmic Flow of VSO}
\label{fig:flow}
\end{figure}

~\autoref{fig:flow} manifests the algorithmic flow of VSO. Firstly, the concerned parameters of the algorithm and the features of the problem are provided as the input to start the execution of VSO. Then, the involved operations as described in~\Autoref{section:operations} are performed successively.

\subsection{A Detailed Analysis on the Search Behavior}\label{section:searching_behavior}
Exploitation and exploration are the two cornerstones of search techniques in solving optimization algorithms. If the exploitation ability is too strong, the algorithm may easily fall into local optima. On the other hand, the algorithm may not be able to converge to any possible solution of a relatively high fitness value in case it solely relies on a very powerful exploration mechanism.

With the novel design of VSO as clearly explained in the previous subsections, it is obvious that VSO combines both advantages of SIs and EAs in order to achieve an excellent balance between exploitation and exploration. The search behavior of the VSO algorithm is summarized as follows.

\subsubsection{Exploitation}
\begin{itemize}
\item As the $critical$ host representing the best solution obtained so far, it has the highest infection rate. Thus, it is more likely to infect $healthy$ hosts to become the $severe$ hosts in the next iteration. To perform such infection operation, the viral RNA of the $critical$ host will be directly replicated to the newly infected host according to~\Autoref{section:infection}. This implies that an increasing number of $healthy$ hosts may acquire the valuable search information of the currently best solution to become $severe$ hosts. On the other hand, since the mutation intensity of the $severe$ hosts will be decreased rapidly as described in~\Autoref{section:mutation}, the $severe$ hosts will conduct neighborhood searches around the locally optimal solution. Overall speaking, this will surely help to enhance the exploitation ability of VSO to improve its solution quality;
\item Each time when a better solution is found, the previous $critical$ host will be automatically downgraded to a $severe$ host to continue its neighborhood search around the previous best solution for a certain duration as seen in~\Autoref{section:infection} before any possible transformation to another host type. Meanwhile, the downgraded host is able to infect other healthy hosts to search this area together.
\end{itemize}

\subsubsection{Exploration}
\begin{itemize}
	\item All $healthy$ hosts of the VSO algorithm perform random exploration to try to find a better solution of the whole search space;
	\item The main role of $mild$ hosts is to improve the exploration capacity of the VSO algorithm. When a $healthy$ host is infected to become a $mild$ host, the viral RNA of the infectious one will not be replicated directly to the $healthy$ host. Instead,  a uni-directional infection mechanism as presented in~\Autoref{section:infection} is performed, that is different from the two-sided crossover operation used in EAs. Moreover, a $mild$ host can always mutate with a higher degree of freedom as guided by the computed intensity. This infection scheme empowers the VSO algorithm with an outstanding exploration ability;
	\item Due to the recovery mechanism, the infected hosts will be recovered and re-initialized from time to time. The recovery mechanism helps to escape from any local minimum for a better exploration;
	\item The imported infection mechanism hybridizes the whole population of the VSO algorithm with another new colony using a totally different searching approach. This may possibly enlarge the search scope of the VSO algorithm for tackling more complex optimization problems. 
\end{itemize}

\begin{figure}[!htb]
	\centerfloat
	\includegraphics[scale=0.42]{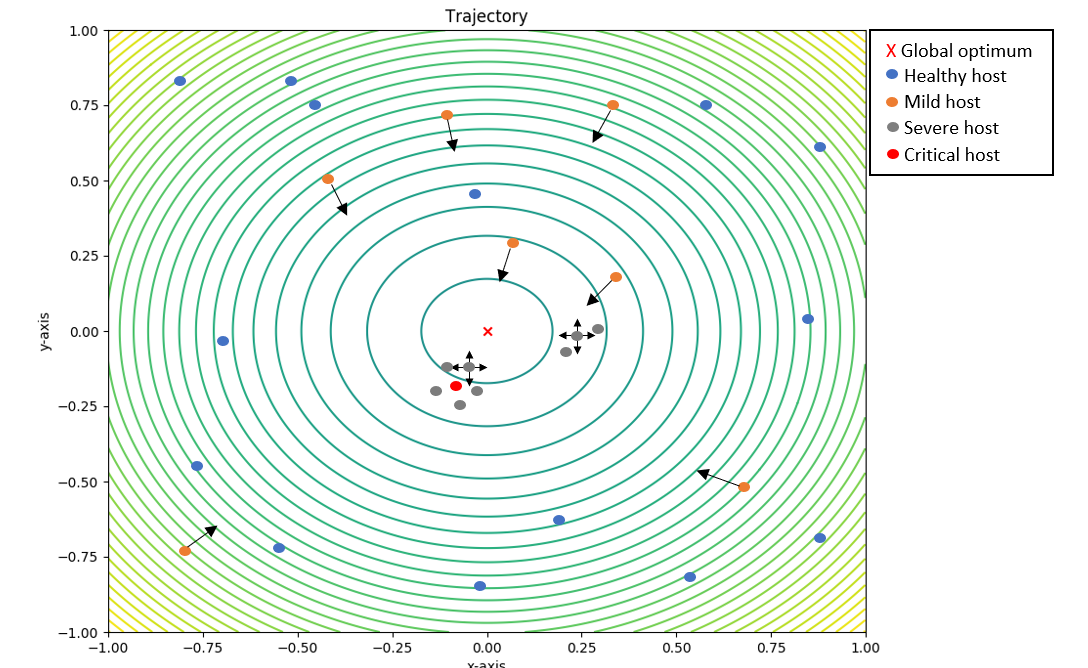}
	\caption{The Searching Pattern of VSO}
	\label{fig:searching_pattern}
\end{figure}

As illustrated in ~\autoref{fig:searching_pattern}, the red cross denotes the globally optimal solution of the specific function while the only red dot represents the $critical$ host as the best solution obtained so far. Clearly, this $critical$ host infect several $severe$ hosts as denoted by gray dots around the central circle to look better solutions whereas the $mild$ hosts as represented by the orange dots will continue to search toward the red dot that is very likely to achieve a near optimal solution.

\subsubsection{Parameters Setting}
Below is a few basic rules for the parameters setting of the VSO algorithm.
\begin{itemize}
	\item The higher the value of $R^M$, $P^C_{H->M}$ or $\gamma$, the better the global search ability of the VSO algorithm, and vice versa; 
	\item On the other hand, the larger the value of $R^C$, $R^S$, $P^C_{H->S}$, $P^S_{H->S}$, or $\alpha$, or the smaller the value of $\delta_s$, the better the local search capability of the VSO algorithm, and vice versa;
	\item  $R$ and $revPercent$ are the conflicting parameters to balance the convergence of the algorithm. $R$ should not be very large, and generally depends on the population size $N_{pop}$ of the VSO algorithm. For instance, $R$ can be set to $1$ for a specific problem with the population size as $50$ to be discussed in the subsequent section;
	\item A larger value of $P_{im}$ may sometimes help to get some quick improvement in solving specific complex optimization problems. Yet for a relatively large value of $P_{im}$, it may also break the good searching patterns. From the empirical observations,  $P_{im} \in (0,0.5]$ is typically a good choice for most benchmark problem sets carefully examined in this work.
\end{itemize}

Because of the diverse searching strategies utilized in the VSO algorithm, the number of parameters is relatively larger than other popular meta-heurisitc algorithms such as GA, PSO, etc.
Yet from the preliminary observations, the performance of the VSO algorithm is relatively robust when only a few of the aforementioned parameters are changed at the same time. Moreover, it is found that the VSO algorithm can flexibly tackle a variety of optimization problems using the same parameter settings without much tuning. As revealed in~\Autoref{section:benchmark_function_evaluations}, the same parameter settings of the VSO algorithm are consistently used in all the following experiments.

\section{Evaluations on Benchmark Functions}\label{section:benchmark_function_evaluations}
To validate both the efficiency and effectiveness of the proposed algorithm, the VSO algorithm is utilized to solve two benchmark function groups including the classical and IEEE CEC 2014 benchmark functions. 

For the classical benchmark functions, a total of $16$ well-known functions given in \autoref{tab:classical_benchmarking_functions} are used. These functions have been well-tested in all kinds of studies of meta-heuristic algorithms in previous research. Among the functions, $F1-F8$ are uni-modal functions while $F9-F16$ are multi-modal functions. Besides, all functions can be scalable from $2$ to $1,000$ dimensions so that the scalability of the concerned algorithms can be investigated. The motivation for testing these classical functions is outlined as follows:
\begin{itemize}
	\item To quickly evaluate the searching capability of VSO when compared to those of other popular meta-heuristic algorithms, especially in terms of the solution quality;
	\item To evaluate the rate of convergence;
	\item To test the reliability of the algorithm;
	\item To investigate the scalability of the algorithm.
\end{itemize}

On the other hand, the IEEE CEC 2014 benchmark functions, as shown in \autoref{tab:CEC_benchmark_functions}, are specially designed for evaluating the performance of meta-heuristic algorithms in the competition of single objective real-parameter numerical optimization problems. The functions ($CEC1-CEC30$)  contain various novel characteristics such as shiftings and rotations, it is much more difficult to solve them than the classical set. Up to our understanding, no algorithm has solved all functions optimally. More details about these functions can be found in~\cite{RN29}. Despite the difficulty, we evaluate the effectiveness and robustness of the VSO algorithm on this set of challenging functions.

In the following experiments, all results are collected on the same computer with the Intel Core i9-7900X CPU running at $3.3\sim4.5$ GHz and $64$ GB of RAM. All algorithms were implemented in $Python3$. ~\autoref{tab:parameters-setting} lists the parameter settings of each concerned algorithm according to the recommended values reported in the literature. Except for the population size, there are totally $11$ unique parameters in VSO as listed in \autoref{tab:parameters-setting}. In fact, for other parameters that are not listed, they can be derived according to the relationships mentioned in~\Autoref{section:operations}. For instance, since  $P^C_{H->S}$ is set to $0.8$, $P^C_{H->M} $is $0.2$. Furthermore, with the imported infection operation, the population size of the main process of the VSO algorithm is consistently set as $30$ while that of the imported infection is $20$. It is worth noting that the parameters of each algorithm remain unchanged in all experiments in order to evaluate the adaptability of the underlying algorithm with the same parameter settings on various problem sets for a fair comparison.

\begin{table*}[!htb]
	\centering
	\footnotesize
	\caption{Parameters Setting}
	\label{tab:parameters-setting}
    
	\begin{tabular}{@{}lll@{}}
		\toprule
		Algorithm & Parameter                                   & Value                               \\ \midrule
		ABC       & Population size                             & 50                                  \\
		& Elite bees num                              & 16                                  \\
		& Onlooker bess num                           & 4                                   \\
		& Patch size                                 & 5                                   \\
		& Patch factor                               & 0.985                               \\
		& Sites num                                  & 3                                   \\
		& Elite sites num                             & 1                                   \\
		CMA-ES    & Population size                             & 4+3*log(dim)                       \\
		& Initial mean                             & 0                                   \\
		& $\sigma$                       & 0.5                                 \\
		DE        & Population size                             & 50                                  \\
		& Crossover rate                             & 0.3                                 \\
		& Differential weight,                        & 0.5                                 \\
		GA        & Population size                             & 50                                  \\
		& Probability of mutation                                  & 0.001                               \\
		& Selection tournsize                         & 3                                   \\
		PSO       & Population size                             & 50                                  \\
		& Inertial weight                             & 0.8                                 \\
		& Cognitive constant                          & 0.5                                 \\
		& Social constant                             & 0.5                                 \\
		SSA       & Population size                             & 50                                  \\
		& $P_a$                                         & 1                                   \\
		& $P_c$                                          & 0.7                                 \\
		& $P_m$                                          & 0.1                                 \\
		VSO       & Population size                             & 50 \\
	    &$R^C$ &  0.8   \\
	    &$R^S$ &  0.3   \\
	    &$R^M$ & 0.3   \\
	    &$P^C_{H->S}$  & 0.8   \\
    	&$P^S_{H->S}$ & 0.5   \\
    	&$\delta_s$    & 0.9   \\
    	&$\alpha$   & 0.1   \\
    	&$\gamma$  & 2     \\ 
    	&$revPercent$ & 0.8   \\
    	&$P_{im}$ & 0.5 \\ 
    	&$H$ & 1 \\
		WOA       & Population size                             & 50                                  \\
		& Initial a                                   & 2                                   \\
		& Probability of Spiral updating     & 0.5                                 \\
		& Constant of shape & 1                                   \\ \bottomrule
	\end{tabular}
\end{table*}
\subsection{Classical Benchmark Functions}\label{section: classical_benchmark_functions_test}
The classical benchmark functions with $30$ dimensions have been widely used for evaluating many meta-heurisitic algorithms like PSO, GA, etc., in many previous studies. In the following evaluation, each function is tested over $31$ runs for each algorithm. The maximum number of iterations in each run is $10^4$. \autoref{tab:classical-func-30} shows the relevant results with the mean as the average value of the fitness values obtained over all runs. The standard deviation of the fitness values is calculated to examine the robustness of the algorithms. Furthermore, the best and worst results are carefully considered. To investigate the computational complexity, the average computational time in CPU seconds is recorded. Finally, two rankings in terms of the averaged fitness values and computational times are listed in order to make more precise and objective comparisons on the different performance measures of the underlying algorithms.

\begin{itemize}
	\item In respect of the uni-modal functions $F1$-$F8$, VSO consistently beats other algorithms in all the rankings. For multi-modal functions $F9$-$F16$, the VSO algorithm gets the first places for $6$ functions as well. More importantly, VSO achieves the exact global optima for all the $12$ functions, i.e. $F1$-$F9$, $F11$, $F13$, and $F15$. The standard deviations are $0$ for all these cases, thus showing the excellent robustness of VSO;
	\item As for other algorithms, we can observe that the performance of CMA-ES and WOA are not bad for the uni-modal functions. Regarding multi-modal ones ($F9$-$F16$), it is clear that CMA-ES goes worse but WOA still works well; 
	\item For both GA and ABC, their performances are not satisfied for multi-modal functions because they may not be good at solving these relatively high dimensional and complex problems; 
	\item Regarding the DE and SSA algorithms, although they acquire very small errors in some functions, they cannot find the exact global optima;
	\item Due to the simple and efficient searching strategies, PSO is very fast. It ranks as the first place in computational time in $9$ cases. Unfortunately, the performance of fitness is worst among all the algorithms.	
\end{itemize}
    
~\autoref{tab:classical-30-summary} shows the summary of classical function evaluations where the average of the rankings in all functions for each algorithm is computed. VSO ranks as the first place with respect to the fitness values whereas it is ranked as the fourth place in terms of the computational time.

\begingroup
\setlength{\LTleft}{-20cm plus -1fill}
\setlength{\LTright}{\LTleft}

\footnotesize{
	\begin{longtable}[]{clrrrrrrrr}
		\caption{Results of Classical Benchmarking Functions}
		\label{tab:classical-func-30}\\
		\hline
		Function &
		Metric &
		\multicolumn{1}{c}{ABC} &
		\multicolumn{1}{c}{CMA-ES} &
		\multicolumn{1}{c}{DE} &
		\multicolumn{1}{c}{GA} &
		\multicolumn{1}{c}{PSO} &
		\multicolumn{1}{c}{SSA} &
		\multicolumn{1}{c}{VSO} &
		\multicolumn{1}{c}{WOA} \\ \hline
		\endhead
		\hline
		\endfoot
		\endlastfoot
		F1 &
		Mean &
		3.18E+06 &
		0.00E+00 &
		3.33E-54 &
		2.97E-11 &
		4.40E+04 &
		6.15E-40 &
		0.00E+00 &
		0.00E+00 \\
		&
		Std &
		3.08E+05 &
		0.00E+00 &
		2.46E-54 &
		1.08E-11 &
		2.57E+04 &
		4.07E-40 &
		0.00E+00 &
		0.00E+00 \\
		&
		Best &
		2.23E+06 &
		0.00E+00 &
		3.29E-55 &
		6.56E-12 &
		1.18E+04 &
		1.57E-40 &
		0.00E+00 &
		0.00E+00 \\
		&
		Worst &
		3.68E+06 &
		0.00E+00 &
		6.12E-54 &
		5.69E-11 &
		1.30E+05 &
		2.19E-39 &
		0.00E+00 &
		0.00E+00 \\
		&
		Time(s) &
		11.31 &
		222.32 &
		72.33 &
		44.29 &
		4.98 &
		5.67 &
		9.49 &
		31.53 \\
		&
		Fitness Rank &
		8 &
		\cellcolor[HTML]{E7E6E6}\textbf{1} &
		4 &
		6 &
		7 &
		5 &
		\cellcolor[HTML]{E7E6E6}\textbf{1} &
		\cellcolor[HTML]{E7E6E6}\textbf{1} \\
		&
		Time Rank &
		4 &
		8 &
		7 &
		6 &
		\cellcolor[HTML]{E7E6E6}\textbf{1} &
		2 &
		3 &
		5 \\
		F2 &
		Mean &
		2.26E-02 &
		0.00E+00 &
		4.57E-53 &
		5.15E-14 &
		2.84E+01 &
		9.53E-44 &
		0.00E+00 &
		0.00E+00 \\
		&
		Std &
		1.22E-01 &
		0.00E+00 &
		4.87E-53 &
		1.26E-29 &
		2.20E+01 &
		7.65E-44 &
		0.00E+00 &
		0.00E+00 \\
		&
		Best &
		3.05E-131 &
		0.00E+00 &
		4.88E-55 &
		5.15E-14 &
		6.73E+00 &
		7.26E-45 &
		0.00E+00 &
		0.00E+00 \\
		&
		Worst &
		6.79E-01 &
		0.00E+00 &
		1.22E-52 &
		5.15E-14 &
		7.42E+01 &
		3.30E-43 &
		0.00E+00 &
		0.00E+00 \\
		&
		Time(s) &
		224.53 &
		498.46 &
		218.26 &
		170.04 &
		155.86 &
		157.80 &
		230.81 &
		190.66 \\
		&
		Fitness Rank &
		7 &
		\cellcolor[HTML]{E7E6E6}\textbf{1} &
		4 &
		6 &
		8 &
		5 &
		\cellcolor[HTML]{E7E6E6}\textbf{1} &
		\cellcolor[HTML]{E7E6E6}\textbf{1} \\
		&
		Time Rank &
		6 &
		8 &
		5 &
		3 &
		\cellcolor[HTML]{E7E6E6}\textbf{1} &
		2 &
		7 &
		4 \\
		F3 &
		Mean &
		1.15E-02 &
		5.25E-284 &
		1.97E-51 &
		3.84E-09 &
		1.38E+05 &
		5.51E-41 &
		0.00E+00 &
		0.00E+00 \\
		&
		Std &
		6.17E-02 &
		0.00E+00 &
		9.36E-52 &
		0.00E+00 &
		1.49E+05 &
		5.80E-41 &
		0.00E+00 &
		0.00E+00 \\
		&
		Best &
		7.31E-128 &
		3.64E-289 &
		9.97E-52 &
		3.84E-09 &
		8.93E+03 &
		9.96E-42 &
		0.00E+00 &
		0.00E+00 \\
		&
		Worst &
		3.44E-01 &
		2.60E-283 &
		3.46E-51 &
		3.84E-09 &
		7.28E+05 &
		2.17E-40 &
		0.00E+00 &
		0.00E+00 \\
		&
		Time(s) &
		108.83 &
		413.26 &
		167.20 &
		78.65 &
		55.84 &
		102.46 &
		152.77 &
		121.10 \\
		&
		Fitness Rank &
		7 &
		3 &
		4 &
		6 &
		8 &
		5 &
		\cellcolor[HTML]{E7E6E6}\textbf{1} &
		\cellcolor[HTML]{E7E6E6}\textbf{1} \\
		&
		Time Rank &
		4 &
		8 &
		7 &
		2 &
		\cellcolor[HTML]{E7E6E6}\textbf{1} &
		3 &
		6 &
		5 \\
		F4 &
		Mean &
		6.08E+01 &
		0.00E+00 &
		1.37E-54 &
		2.30E-13 &
		1.50E+01 &
		1.09E-41 &
		0.00E+00 &
		5.32E-233 \\
		&
		Std &
		4.15E+00 &
		0.00E+00 &
		1.40E-54 &
		7.78E-14 &
		2.53E+00 &
		7.21E-42 &
		0.00E+00 &
		0.00E+00 \\
		&
		Best &
		4.44E+01 &
		0.00E+00 &
		4.14E-55 &
		1.17E-13 &
		9.14E+00 &
		2.72E-42 &
		0.00E+00 &
		1.77E-269 \\
		&
		Worst &
		6.56E+01 &
		0.00E+00 &
		4.15E-54 &
		4.81E-13 &
		2.13E+01 &
		2.81E-41 &
		0.00E+00 &
		1.60E-231 \\
		&
		Time(s) &
		11.67 &
		229.63 &
		71.92 &
		43.76 &
		5.06 &
		8.95 &
		9.68 &
		30.90 \\
		&
		Fitness Rank &
		8 &
		\cellcolor[HTML]{E7E6E6}\textbf{1} &
		4 &
		6 &
		7 &
		5 &
		\cellcolor[HTML]{E7E6E6}\textbf{1} &
		3 \\
		&
		Time Rank &
		4 &
		8 &
		7 &
		6 &
		\cellcolor[HTML]{E7E6E6}\textbf{1} &
		2 &
		3 &
		5 \\
		F5 &
		Mean &
		3.95E-130 &
		0.00E+00 &
		1.04E-52 &
		6.77E-13 &
		1.27E+01 &
		4.09E-43 &
		0.00E+00 &
		0.00E+00 \\
		&
		Std &
		1.28E-130 &
		0.00E+00 &
		1.21E-52 &
		2.02E-28 &
		6.41E+00 &
		3.27E-43 &
		0.00E+00 &
		0.00E+00 \\
		&
		Best &
		1.76E-130 &
		0.00E+00 &
		1.22E-53 &
		6.77E-13 &
		5.29E+00 &
		6.17E-44 &
		0.00E+00 &
		0.00E+00 \\
		&
		Worst &
		6.72E-130 &
		0.00E+00 &
		3.30E-52 &
		6.77E-13 &
		2.97E+01 &
		1.09E-42 &
		0.00E+00 &
		0.00E+00 \\
		&
		Time(s) &
		90.23 &
		328.04 &
		128.58 &
		97.28 &
		61.20 &
		38.05 &
		101.42 &
		82.87 \\
		&
		Fitness Rank &
		4 &
		\cellcolor[HTML]{E7E6E6}\textbf{1} &
		5 &
		7 &
		8 &
		6 &
		\cellcolor[HTML]{E7E6E6}\textbf{1} &
		\cellcolor[HTML]{E7E6E6}\textbf{1} \\
		&
		Time Rank &
		4 &
		8 &
		7 &
		5 &
		2 &
		\cellcolor[HTML]{E7E6E6}\textbf{1} &
		6 &
		3 \\
		F6 &
		Mean &
		1.03E-24 &
		1.04E-57 &
		1.79E-15 &
		8.88E-16 &
		9.92E-08 &
		2.15E-116 &
		0.00E+00 &
		0.00E+00 \\
		&
		Std &
		1.93E-24 &
		1.16E-57 &
		3.58E-15 &
		1.97E-31 &
		1.62E-07 &
		1.43E-116 &
		0.00E+00 &
		0.00E+00 \\
		&
		Best &
		8.13E-28 &
		1.40E-60 &
		1.40E-34 &
		8.88E-16 &
		1.43E-09 &
		5.82E-117 &
		0.00E+00 &
		0.00E+00 \\
		&
		Worst &
		9.12E-24 &
		3.17E-57 &
		8.95E-15 &
		8.88E-16 &
		7.92E-07 &
		7.00E-116 &
		0.00E+00 &
		0.00E+00 \\
		&
		Time(s) &
		73.73 &
		398.39 &
		148.99 &
		106.30 &
		83.30 &
		87.26 &
		130.74 &
		85.20 \\
		&
		Fitness Rank &
		5 &
		4 &
		7 &
		6 &
		8 &
		3 &
		\cellcolor[HTML]{E7E6E6}\textbf{1} &
		\cellcolor[HTML]{E7E6E6}\textbf{1} \\
		&
		Time Rank &
		\cellcolor[HTML]{E7E6E6}\textbf{1} &
		8 &
		7 &
		5 &
		2 &
		4 &
		6 &
		3 \\
		F7 &
		Mean &
		1.95E+02 &
		1.2e-322 &
		3.33E-31 &
		6.05E+00 &
		1.31E+02 &
		5.26E-03 &
		0.00E+00 &
		1.04E-20 \\
		&
		Std &
		3.99E+01 &
		0.00E+00 &
		3.36E-31 &
		2.54E+00 &
		6.93E+01 &
		2.43E-03 &
		0.00E+00 &
		5.59E-20 \\
		&
		Best &
		1.20E+02 &
		0.00E+00 &
		6.92E-33 &
		2.33E+00 &
		2.71E+01 &
		1.84E-03 &
		0.00E+00 &
		1.29E-60 \\
		&
		Worst &
		2.74E+02 &
		5.93e-322 &
		9.46E-31 &
		1.22E+01 &
		3.12E+02 &
		1.35E-02 &
		0.00E+00 &
		3.12E-19 \\
		&
		Time(s) &
		60.39 &
		368.70 &
		137.53 &
		63.75 &
		40.19 &
		69.38 &
		115.98 &
		53.24 \\
		&
		Fitness Rank &
		8 &
		2 &
		3 &
		6 &
		7 &
		5 &
		\cellcolor[HTML]{E7E6E6}\textbf{1} &
		4 \\
		&
		Time Rank &
		3 &
		8 &
		7 &
		4 &
		\cellcolor[HTML]{E7E6E6}\textbf{1} &
		5 &
		6 &
		2 \\
		F8 &
		Mean &
		2.60E+04 &
		3.14E-128 &
		1.87E-34 &
		1.83E+01 &
		1.63E+04 &
		6.07E+01 &
		0.00E+00 &
		4.14E-67 \\
		&
		Std &
		5.04E+03 &
		5.31E-128 &
		2.55E-34 &
		8.94E+00 &
		7.08E+03 &
		1.65E+01 &
		0.00E+00 &
		2.23E-66 \\
		&
		Best &
		1.43E+04 &
		6.57E-258 &
		1.50E-35 &
		3.12E+00 &
		3.63E+03 &
		2.65E+01 &
		0.00E+00 &
		1.32E-245 \\
		&
		Worst &
		3.54E+04 &
		1.37E-127 &
		6.91E-34 &
		3.95E+01 &
		3.12E+04 &
		1.02E+02 &
		0.00E+00 &
		1.24E-65 \\
		&
		Time(s) &
		378.85 &
		534.32 &
		321.64 &
		298.88 &
		257.02 &
		264.40 &
		271.99 &
		175.60 \\
		&
		Fitness Rank &
		8 &
		2 &
		4 &
		5 &
		7 &
		6 &
		\cellcolor[HTML]{E7E6E6}\textbf{1} &
		3 \\
		&
		Time Rank &
		7 &
		8 &
		6 &
		5 &
		2 &
		3 &
		4 &
		\cellcolor[HTML]{E7E6E6}\textbf{1} \\
		F9 &
		Mean &
		1.24E+01 &
		4.36E+01 &
		2.49E+01 &
		3.80E+01 &
		8.67E+01 &
		2.25E-13 &
		0.00E+00 &
		0.00E+00 \\
		&
		Std &
		3.30E+00 &
		9.55E+00 &
		1.89E+00 &
		9.42E+00 &
		1.78E+01 &
		7.95E-13 &
		0.00E+00 &
		0.00E+00 \\
		&
		Best &
		5.97E+00 &
		3.18E+01 &
		2.19E+01 &
		1.79E+01 &
		4.21E+01 &
		0.00E+00 &
		0.00E+00 &
		0.00E+00 \\
		&
		Worst &
		1.89E+01 &
		5.97E+01 &
		2.69E+01 &
		5.79E+01 &
		1.30E+02 &
		4.23E-12 &
		0.00E+00 &
		0.00E+00 \\
		&
		Time(s) &
		18.28 &
		197.22 &
		76.80 &
		44.99 &
		9.26 &
		13.10 &
		12.63 &
		26.09 \\
		&
		Fitness Rank &
		4 &
		7 &
		5 &
		6 &
		8 &
		3 &
		\cellcolor[HTML]{E7E6E6}\textbf{1} &
		\cellcolor[HTML]{E7E6E6}\textbf{1} \\
		&
		Time Rank &
		4 &
		8 &
		7 &
		6 &
		\cellcolor[HTML]{E7E6E6}\textbf{1} &
		3 &
		2 &
		5 \\
		F10 &
		Mean &
		1.72E+01 &
		2.00E+01 &
		8.97E-15 &
		1.27E-07 &
		9.47E+00 &
		7.90E-15 &
		4.44E-16 &
		4.44E-16 \\
		&
		Std &
		8.01E-01 &
		1.17E-02 &
		2.84E-15 &
		1.26E-08 &
		1.23E+00 &
		1.41E-15 &
		0.00E+00 &
		0.00E+00 \\
		&
		Best &
		1.48E+01 &
		2.00E+01 &
		7.55E-15 &
		1.19E-07 &
		7.36E+00 &
		7.55E-15 &
		4.44E-16 &
		4.44E-16 \\
		&
		Worst &
		1.84E+01 &
		2.00E+01 &
		1.47E-14 &
		1.71E-07 &
		1.20E+01 &
		1.47E-14 &
		4.44E-16 &
		4.44E-16 \\
		&
		Time(s) &
		21.05 &
		112.15 &
		83.13 &
		47.54 &
		15.59 &
		18.52 &
		23.63 &
		40.88 \\
		&
		Fitness Rank &
		7 &
		8 &
		4 &
		5 &
		6 &
		3 &
		\cellcolor[HTML]{E7E6E6}\textbf{1} &
		\cellcolor[HTML]{E7E6E6}\textbf{1} \\
		&
		Time Rank &
		3 &
		8 &
		7 &
		6 &
		\cellcolor[HTML]{E7E6E6}\textbf{1} &
		2 &
		4 &
		5 \\
		F11 &
		Mean &
		7.75E+00 &
		1.97E-03 &
		0.00E+00 &
		6.93E-02 &
		1.10E+00 &
		0.00E+00 &
		0.00E+00 &
		0.00E+00 \\
		&
		Std &
		7.05E-01 &
		3.94E-03 &
		0.00E+00 &
		1.37E-01 &
		7.31E-02 &
		0.00E+00 &
		0.00E+00 &
		0.00E+00 \\
		&
		Best &
		6.46E+00 &
		0.00E+00 &
		0.00E+00 &
		5.77E-15 &
		9.54E-01 &
		0.00E+00 &
		0.00E+00 &
		0.00E+00 \\
		&
		Worst &
		9.02E+00 &
		9.86E-03 &
		0.00E+00 &
		7.73E-01 &
		1.33E+00 &
		0.00E+00 &
		0.00E+00 &
		0.00E+00 \\
		&
		Time(s) &
		16.77 &
		206.10 &
		80.82 &
		52.69 &
		8.78 &
		18.81 &
		32.01 &
		41.97 \\
		&
		Fitness Rank &
		8 &
		5 &
		\cellcolor[HTML]{E7E6E6}\textbf{1} &
		6 &
		7 &
		\cellcolor[HTML]{E7E6E6}\textbf{1} &
		\cellcolor[HTML]{E7E6E6}\textbf{1} &
		\cellcolor[HTML]{E7E6E6}\textbf{1} \\
		&
		Time Rank &
		2 &
		8 &
		7 &
		6 &
		\cellcolor[HTML]{E7E6E6}\textbf{1} &
		3 &
		4 &
		5 \\
		F12 &
		Mean &
		-1.04E+03 &
		-8.92E+02 &
		-7.51E+02 &
		-1.17E+03 &
		-1.01E+03 &
		-5.52E+02 &
		-1.17E+03 &
		-1.17E+03 \\
		&
		Std &
		2.75E+01 &
		1.07E+02 &
		0.00E+00 &
		2.12E-13 &
		3.76E+01 &
		4.16E+01 &
		6.48E+00 &
		4.49E-05 \\
		&
		Best &
		-1.09E+03 &
		-1.01E+03 &
		-7.51E+02 &
		-1.17E+03 &
		-1.09E+03 &
		-6.32E+02 &
		-1.17E+03 &
		-1.17E+03 \\
		&
		Worst &
		-9.91E+02 &
		-7.51E+02 &
		-7.51E+02 &
		-1.17E+03 &
		-9.47E+02 &
		-4.74E+02 &
		-1.16E+03 &
		-1.17E+03 \\
		&
		Time(s) &
		183.06 &
		429.66 &
		192.95 &
		93.44 &
		72.23 &
		130.79 &
		93.04 &
		152.13 \\
		&
		Fitness Rank &
		4 &
		6 &
		7 &
		\cellcolor[HTML]{E7E6E6}\textbf{1} &
		5 &
		8 &
		3 &
		2 \\
		&
		Time Rank &
		6 &
		8 &
		7 &
		3 &
		\cellcolor[HTML]{E7E6E6}\textbf{1} &
		4 &
		2 &
		5 \\
		F13 &
		Mean &
		0.00E+00 &
		0.00E+00 &
		1.92E-150 &
		1.42E-13 &
		4.05E-06 &
		2.08E-106 &
		0.00E+00 &
		0.00E+00 \\
		&
		Std &
		0.00E+00 &
		0.00E+00 &
		3.77E-150 &
		1.88E-13 &
		4.15E-06 &
		2.45E-106 &
		0.00E+00 &
		0.00E+00 \\
		&
		Best &
		0.00E+00 &
		0.00E+00 &
		7.65E-158 &
		2.22E-22 &
		1.05E-06 &
		3.40E-108 &
		0.00E+00 &
		0.00E+00 \\
		&
		Worst &
		0.00E+00 &
		0.00E+00 &
		9.46E-150 &
		4.71E-13 &
		1.23E-05 &
		5.73E-106 &
		0.00E+00 &
		0.00E+00 \\
		&
		Time(s) &
		16.77 &
		98.05 &
		47.45 &
		39.50 &
		4.21 &
		13.32 &
		22.69 &
		33.79 \\
		&
		Fitness Rank &
		\cellcolor[HTML]{E7E6E6}\textbf{1} &
		\cellcolor[HTML]{E7E6E6}\textbf{1} &
		5 &
		7 &
		8 &
		6 &
		\cellcolor[HTML]{E7E6E6}\textbf{1} &
		\cellcolor[HTML]{E7E6E6}\textbf{1} \\
		&
		Time Rank &
		3 &
		8 &
		7 &
		6 &
		\cellcolor[HTML]{E7E6E6}\textbf{1} &
		2 &
		4 &
		5 \\
		F14 &
		Mean &
		8.54E-12 &
		1.71E-11 &
		3.51E-12 &
		7.38E-12 &
		2.50E-11 &
		5.17E-12 &
		3.51E-12 &
		3.51E-12 \\
		&
		Std &
		9.04E-13 &
		1.33E-12 &
		2.54E-27 &
		8.82E-13 &
		3.03E-12 &
		3.99E-13 &
		4.96E-26 &
		2.69E-17 \\
		&
		Best &
		7.17E-12 &
		1.46E-11 &
		3.51E-12 &
		5.70E-12 &
		1.70E-11 &
		4.34E-12 &
		3.51E-12 &
		3.51E-12 \\
		&
		Worst &
		1.12E-11 &
		1.82E-11 &
		3.51E-12 &
		9.10E-12 &
		3.00E-11 &
		5.89E-12 &
		3.51E-12 &
		3.51E-12 \\
		&
		Time(s) &
		159.41 &
		334.09 &
		180.82 &
		154.80 &
		118.05 &
		123.38 &
		87.64 &
		79.20 \\
		&
		Fitness Rank &
		6 &
		7 &
		\cellcolor[HTML]{E7E6E6}\textbf{1} &
		5 &
		8 &
		4 &
		2 &
		3 \\
		&
		Time Rank &
		6 &
		8 &
		7 &
		5 &
		3 &
		4 &
		2 &
		\cellcolor[HTML]{E7E6E6}\textbf{1} \\
		F15 &
		Mean &
		2.24E-14 &
		2.00E-12 &
		3.60E-18 &
		1.75E-06 &
		7.26E+00 &
		2.24E-03 &
		0.00E+00 &
		0.00E+00 \\
		&
		Std &
		7.69E-15 &
		2.64E-13 &
		2.56E-18 &
		3.97E-07 &
		4.70E+00 &
		5.30E-04 &
		0.00E+00 &
		0.00E+00 \\
		&
		Best &
		9.66E-15 &
		1.53E-12 &
		6.39E-19 &
		9.84E-07 &
		1.08E+00 &
		1.22E-03 &
		0.00E+00 &
		0.00E+00 \\
		&
		Worst &
		4.17E-14 &
		2.25E-12 &
		7.65E-18 &
		2.51E-06 &
		2.07E+01 &
		3.48E-03 &
		0.00E+00 &
		0.00E+00 \\
		&
		Time(s) &
		123.42 &
		362.59 &
		144.51 &
		112.47 &
		82.66 &
		77.79 &
		121.00 &
		112.19 \\
		&
		Fitness Rank &
		4 &
		5 &
		3 &
		6 &
		8 &
		7 &
		\cellcolor[HTML]{E7E6E6}\textbf{1} &
		\cellcolor[HTML]{E7E6E6}\textbf{1} \\
		&
		Time Rank &
		6 &
		8 &
		7 &
		4 &
		2 &
		\cellcolor[HTML]{E7E6E6}\textbf{1} &
		5 &
		3 \\
		F16 &
		Mean &
		-2.77E+01 &
		-1.86E+01 &
		-2.40E+01 &
		-2.62E+01 &
		-1.73E+01 &
		-4.33E+00 &
		-2.88E+01 &
		-1.25E+01 \\
		&
		Std &
		5.57E-01 &
		1.60E+00 &
		1.25E+00 &
		6.03E-01 &
		2.04E+00 &
		6.37E-01 &
		1.91E-01 &
		1.86E+00 \\
		&
		Best &
		-2.86E+01 &
		-2.03E+01 &
		-2.64E+01 &
		-2.73E+01 &
		-2.05E+01 &
		-6.17E+00 &
		-2.91E+01 &
		-1.57E+01 \\
		&
		Worst &
		-2.66E+01 &
		-1.56E+01 &
		-2.29E+01 &
		-2.46E+01 &
		-1.18E+01 &
		-3.15E+00 &
		-2.85E+01 &
		-7.73E+00 \\
		&
		Time(s) &
		262.23 &
		569.71 &
		248.92 &
		219.29 &
		125.46 &
		122.20 &
		254.45 &
		116.78 \\
		&
		Fitness Rank &
		2 &
		5 &
		4 &
		3 &
		6 &
		8 &
		\cellcolor[HTML]{E7E6E6}\textbf{1} &
		7 \\
		&
		Time Rank &
		7 &
		8 &
		5 &
		4 &
		3 &
		2 &
		6 &
		\cellcolor[HTML]{E7E6E6}\textbf{1} \\ \hline
	\end{longtable}
}

\endgroup

\begin{table*}[h]
	\footnotesize
	\centering
	\caption{Summary of Classical Benchmarking Function Evaluations}
	\label{tab:classical-30-summary}
    \begin{tabular}{@{}lcccc@{}}
	\toprule
	Algorithm &
	\begin{tabular}[c]{@{}c@{}}Avg Fitness\\ Rank\end{tabular} &
	\begin{tabular}[c]{@{}c@{}}Avg Time\\ Rank\end{tabular} &
	\begin{tabular}[c]{@{}c@{}}Overall Fitness\\ Rank\end{tabular} &
	\begin{tabular}[c]{@{}c@{}}Overall Time\\ Rank\end{tabular} \\ \midrule
	ABC    & 5.69 & 4.38 & 7                                  & 4                                  \\
	CMA-ES & 3.69 & 8.00 & 3                                  & 8                                  \\
	DE     & 4.06 & 6.69 & 4                                  & 7                                  \\
	GA     & 5.44 & 4.75 & 6                                  & 6                                  \\
	PSO    & 7.25 & 1.50 & 8                                  & \cellcolor[HTML]{E7E6E6}\textbf{1} \\
	SSA    & 5.00 & 2.69 & 5                                  & 2                                  \\
	VSO    & 1.19 & 4.38 & \cellcolor[HTML]{E7E6E6}\textbf{1} & 4                                  \\
	WOA    & 2.00 & 3.63 & 2                                  & 3                                  \\ \bottomrule
    \end{tabular}
\end{table*}
    
\subsection{CEC benchmark Functions Test}
Following the CEC 2014 recommendation~\cite{RN29}, the dimension of problems is selected as $30$ as well. Function $CEC1$-$CEC3$ are uni-modal functions with rotations. $CEC4$-$CEC16$ are simple multi-modal functions but with various shiftings and rotations. $CEC17$-$CEC22$ are hybrid functions while $CEC23$-$CEC30$ are the composition functions. 
As the global optimum of each function is different (from $100$ to $3000$), the fitness result is converted to the error as calculated in~\eqref{eq:eq13} to make the comparison more straightforward. In other words, when the result gets closer to $0$, it implies that the best solution obtained by the algorithm is closer to the global optimum of the corresponding function.

\begin{equation}\label{eq:eq13}
	fitness = f(x)-f(x^*)
\end{equation}
where $x$ is the best solution obtained by the algorithm while $f(x^*)$ is the real global optimum of the function. 

~\autoref{tab:cec-results} reports the results over $31$ independent runs on each function for each algorithm. A few observations are specified as follows.

\begin{itemize}
	\item For the uni-modal functions $CEC1$-$CEC3$, no algorithm is dominated. Because of the complicated rotation, the errors for $CEC1$ are huge for all algorithms. Only GA and VSO perform relatively better. Both errors are on the same order of magnitude, i.e. $10^6$.  In the case of $CEC2$, it is similar to $CEC1$, only SSA and VSO are on the smallest order ($10^2$) of magnitude. Meanwhile, the best metric of all runs for VSO is $0$, which demonstrates that only VSO has once achieved the exact global optima. Interestingly, WOA performs very well in the classical uni-modal functions but does not work in these complicated uni-modal cases. It is remarkable that only VSO is able to achieve the global optimum exactly (with fitness error $0$) in $CEC3$;
	\item For multi-modal functions $CEC4$-$CEC16$, it is obvious that the VSO algorithm attains a better performance than those of other algorithms. In terms of the mean of fitness, VSO ranks the best for over half of the functions, including the $CEC6$, $CEC9$, $CEC10$, $CEC13$, $CEC14$, $CEC15$ and $CEC16$. SSA followed by VSO acquires the best performance of fitness in $4$ functions;
	\item For $6$ hybrid functions $CEC17$-$CEC22$, the VSO algorithm achieves the best performance in all functions except for $CEC18$ where it ranks the second place;
	\item For the composition functions of $CEC23$-$CEC30$, VSO outperforms all compared algorithms in the first $6$ functions, i.e. $CEC23$-$CEC28$. It can also be observed that WOA obtains the best performance in $CEC23$ and $CEC25$. This verifies the outstanding optimizing capacity of VSO on such complex functions.
\end{itemize}

~\autoref{tab:cec-summary} indicates that VSO generally outperforms all other algorithms in terms of the fitness values.In addition, VSO with the imported infection operation powered by DE works well here yet the performance of the standalone DE is the worst of all.

\clearpage
\begingroup
\setlength{\LTleft}{-20cm plus -1fill}
\setlength{\LTright}{\LTleft}

{\footnotesize
\begin{longtable}[!htb]{clrrrrrrrr}
	\caption{Results of CEC Benchmarking Functions}
	\label{tab:cec-results}\\
	\toprule
	{\color[HTML]{000000} Function} &
	{\color[HTML]{000000} Metric} &
	\multicolumn{1}{c}{{\color[HTML]{000000} ABC}} &
	\multicolumn{1}{c}{{\color[HTML]{000000} CMA-ES}} &
	\multicolumn{1}{c}{{\color[HTML]{000000} DE}} &
	\multicolumn{1}{c}{{\color[HTML]{000000} GA}} &
	\multicolumn{1}{c}{{\color[HTML]{000000} PSO}} &
	\multicolumn{1}{c}{{\color[HTML]{000000} SSA}} &
	\multicolumn{1}{c}{{\color[HTML]{000000} VSO}} &
	\multicolumn{1}{c}{{\color[HTML]{000000} WOA}} \\* \midrule
	\endhead
	\bottomrule
	\endfoot
	\endlastfoot
	{\color[HTML]{000000} CEC1} &
	{\color[HTML]{000000} Mean} &
	{\color[HTML]{000000} 5.15E+08} &
	{\color[HTML]{000000} 5.83E+07} &
	{\color[HTML]{000000} 2.36E+09} &
	{\color[HTML]{000000} 2.07E+06} &
	{\color[HTML]{000000} 2.05E+08} &
	{\color[HTML]{000000} 6.49E+06} &
	{\color[HTML]{000000} 3.17E+06} &
	{\color[HTML]{000000} 3.67E+08} \\
	{\color[HTML]{000000} } &
	{\color[HTML]{000000} Std} &
	{\color[HTML]{000000} 1.39E+08} &
	{\color[HTML]{000000} 2.90E+07} &
	{\color[HTML]{000000} 4.92E-07} &
	{\color[HTML]{000000} 1.35E+06} &
	{\color[HTML]{000000} 2.53E+08} &
	{\color[HTML]{000000} 1.68E+06} &
	{\color[HTML]{000000} 1.28E+06} &
	{\color[HTML]{000000} 1.85E+08} \\
	{\color[HTML]{000000} } &
	{\color[HTML]{000000} Best} &
	{\color[HTML]{000000} 1.61E+08} &
	{\color[HTML]{000000} 9.84E+06} &
	{\color[HTML]{000000} 2.36E+09} &
	{\color[HTML]{000000} 3.56E+05} &
	{\color[HTML]{000000} 3.31E+07} &
	{\color[HTML]{000000} 2.68E+06} &
	{\color[HTML]{000000} 1.21E+06} &
	{\color[HTML]{000000} 1.11E+08} \\
	{\color[HTML]{000000} } &
	{\color[HTML]{000000} Worst} &
	{\color[HTML]{000000} 7.82E+08} &
	{\color[HTML]{000000} 1.49E+08} &
	{\color[HTML]{000000} 2.36E+09} &
	{\color[HTML]{000000} 5.79E+06} &
	{\color[HTML]{000000} 1.40E+09} &
	{\color[HTML]{000000} 1.02E+07} &
	{\color[HTML]{000000} 6.67E+06} &
	{\color[HTML]{000000} 8.87E+08} \\
	{\color[HTML]{000000} } &
	{\color[HTML]{000000} Time(s)} &
	{\color[HTML]{000000} 7.15} &
	{\color[HTML]{000000} 1568.44} &
	{\color[HTML]{000000} 66.78} &
	{\color[HTML]{000000} 47.22} &
	{\color[HTML]{000000} 3.69} &
	{\color[HTML]{000000} 5.90} &
	{\color[HTML]{000000} 10.40} &
	{\color[HTML]{000000} 14.70} \\
\bottomrule
\end{longtable}
}

\endgroup
\clearpage
\begin{table}[!htb]
	\footnotesize
	\centering
	\caption{Summary of Evaluations for CEC Benchmarking Functions}
	\label{tab:cec-summary}
	%
} \\ \midrule
		{\color[HTML]{000000} ABC} &
		{\color[HTML]{000000} 6.07} &
		{\color[HTML]{000000} 3.23} &
		{\color[HTML]{000000} 7} &
		{\color[HTML]{000000} 3} \\
		{\color[HTML]{000000} CMA-ES} &
		{\color[HTML]{000000} 3.93} &
		{\color[HTML]{000000} 7.93} &
		{\color[HTML]{000000} 4} &
		{\color[HTML]{000000} 8} \\
		{\color[HTML]{000000} DE} &
		{\color[HTML]{000000} 6.60} &
		{\color[HTML]{000000} 6.73} &
		{\color[HTML]{000000} 8} &
		{\color[HTML]{000000} 7} \\
		{\color[HTML]{000000} GA} &
		{\color[HTML]{000000} 3.83} &
		{\color[HTML]{000000} 5.77} &
		{\color[HTML]{000000} 3} &
		{\color[HTML]{000000} 6} \\
		{\color[HTML]{000000} PSO} &
		{\color[HTML]{000000} 5.37} &
		{\color[HTML]{000000} 1.20} &
		{\color[HTML]{000000} 5} &
		\cellcolor[HTML]{E7E6E6}{\color[HTML]{000000} \textbf{1}} \\
		{\color[HTML]{000000} SSA} &
		{\color[HTML]{000000} 2.63} &
		{\color[HTML]{000000} 2.03} &
		{\color[HTML]{000000} 2} &
		{\color[HTML]{000000} 2} \\
		{\color[HTML]{000000} VSO} &
		{\color[HTML]{000000} 1.67} &
		{\color[HTML]{000000} 4.03} &
		\cellcolor[HTML]{E7E6E6}{\color[HTML]{000000} \textbf{1}} &
		{\color[HTML]{000000} 4} \\
		{\color[HTML]{000000} WOA} &
		{\color[HTML]{000000} 5.83} &
		{\color[HTML]{000000} 5.07} &
		{\color[HTML]{000000} 6} &
		{\color[HTML]{000000} 5} \\ \bottomrule
	\end{tabular}
\end{table}
\subsection{Convergence Test}
In addition to the solution quality, we are also interested in the rate of convergence. Therefore, the convergence test on those classical benchmark functions is conducted.

\begin{figure*}[!htb]
	\centerfloat
	\subfigure[]{
		\begin{minipage}[t]{0.34\linewidth}
			\centering
			\includegraphics[width=2.0in]{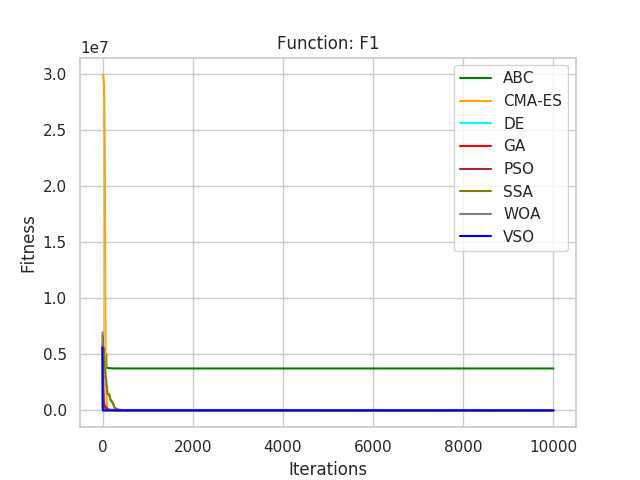}
		\end{minipage}%
	}%
	\subfigure[]{
		\begin{minipage}[t]{0.34\linewidth}
			\centering
			\includegraphics[width=2.0in]{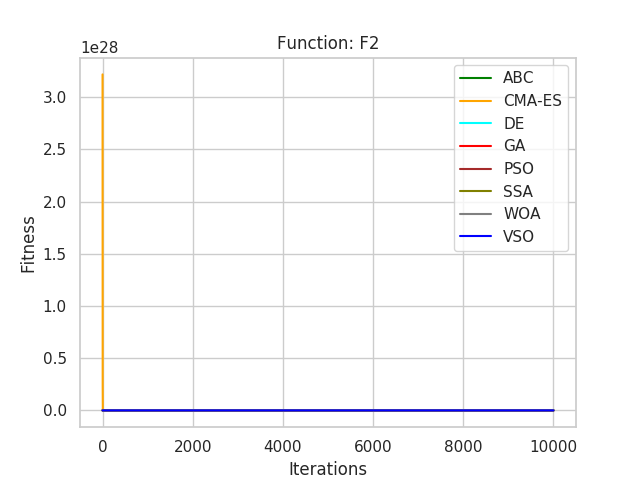}
			
		\end{minipage}%
	}%
	\subfigure[]{
		\begin{minipage}[t]{0.34\linewidth}
			\centering
			\includegraphics[width=2.0in]{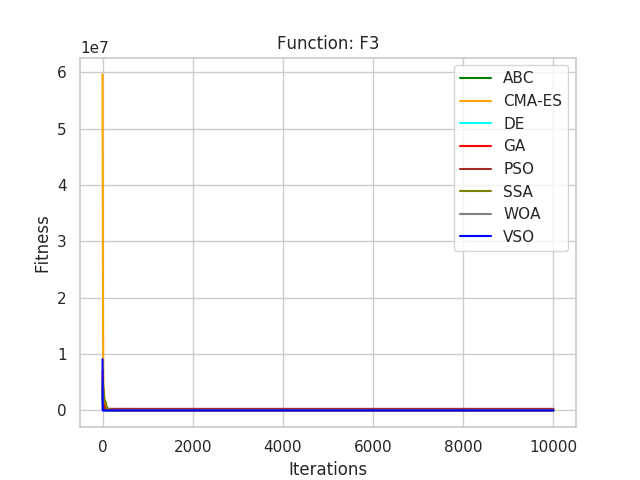}
			
		\end{minipage}
	}%
	\subfigure[]{
		\begin{minipage}[t]{0.34\linewidth}
			\centering
			\includegraphics[width=2.0in]{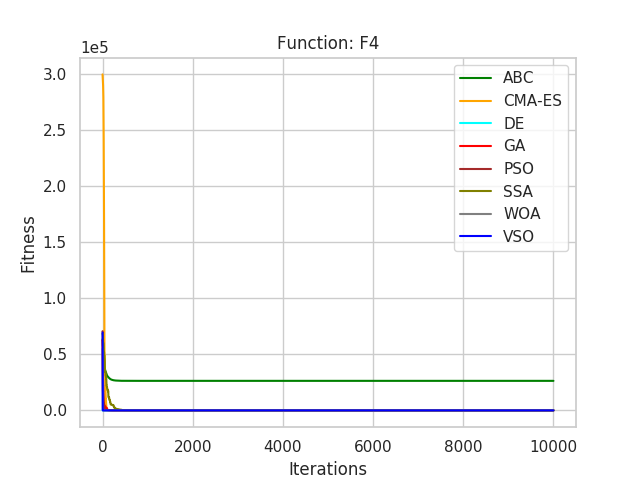}
		\end{minipage}%
	}%

	\subfigure[]{
		\begin{minipage}[t]{0.34\linewidth}
			\centering
			\includegraphics[width=2.0in]{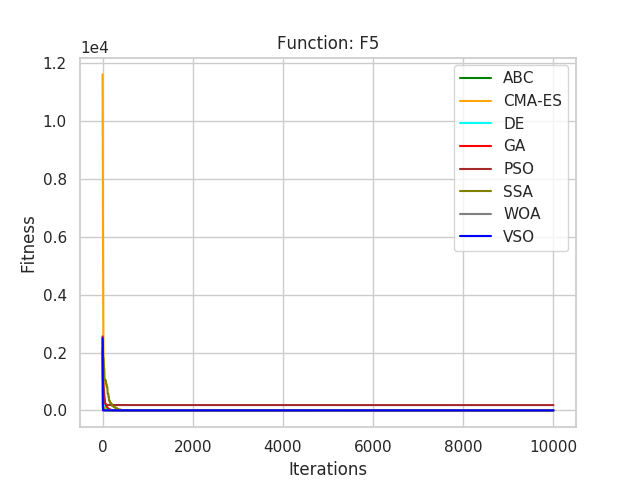}
			
		\end{minipage}%
	}%
	\subfigure[]{
		\begin{minipage}[t]{0.34\linewidth}
			\centering
			\includegraphics[width=2.0in]{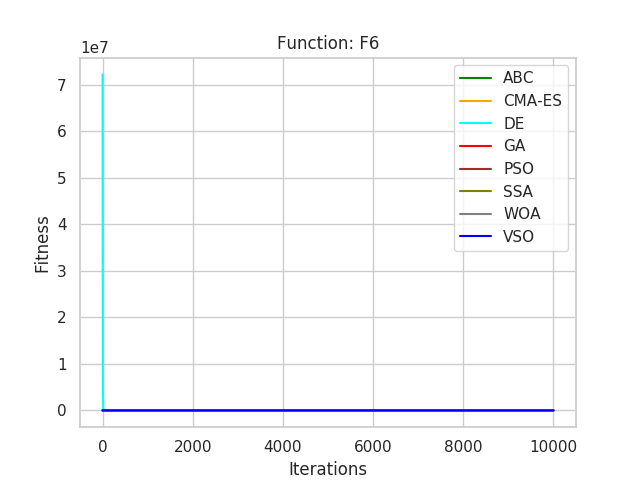}
			
		\end{minipage}
	}%
	\subfigure[]{
		\begin{minipage}[t]{0.34\linewidth}
			\centering
			\includegraphics[width=2.0in]{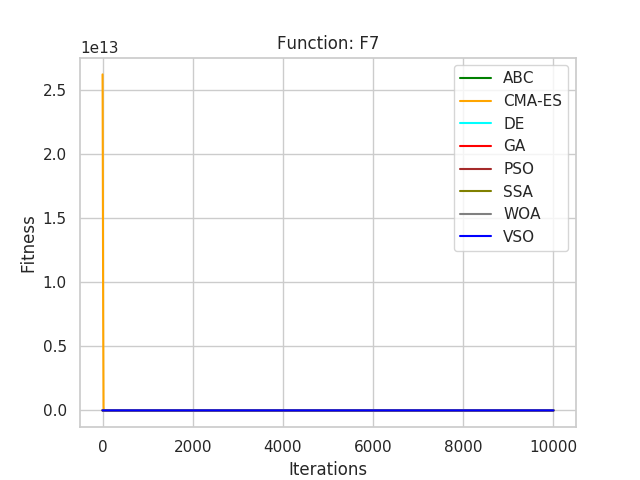}
		\end{minipage}%
	}%
	\subfigure[]{
		\begin{minipage}[t]{0.34\linewidth}
			\centering
			\includegraphics[width=2.0in]{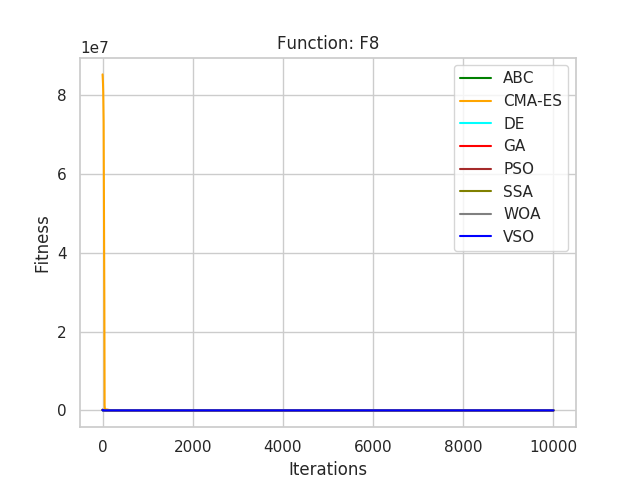}
			
		\end{minipage}%
	}%

	\subfigure[]{
		\begin{minipage}[t]{0.34\linewidth}
			\centering
			\includegraphics[width=2.0in]{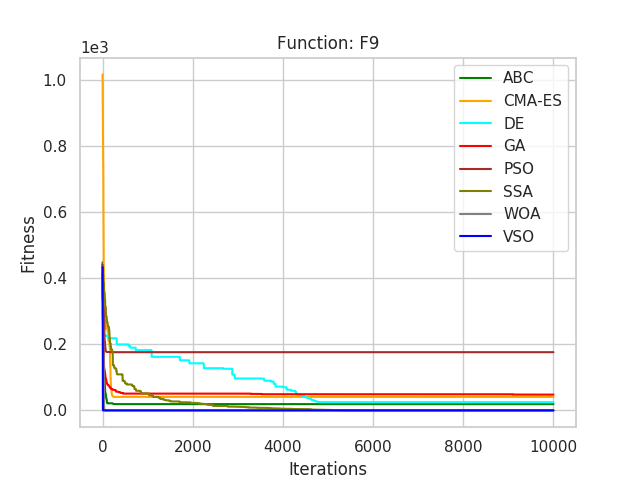}
			
		\end{minipage}
	}%
	\subfigure[]{
		\begin{minipage}[t]{0.34\linewidth}
			\centering
			\includegraphics[width=2.0in]{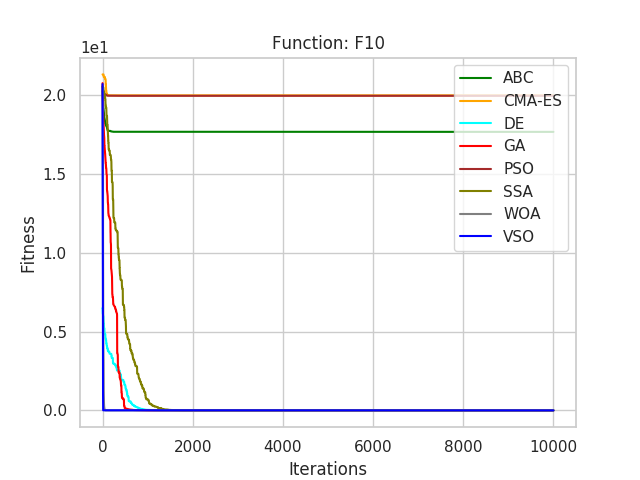}
			
		\end{minipage}
	}%
	\subfigure[]{
		\begin{minipage}[t]{0.34\linewidth}
			\centering
			\includegraphics[width=2.0in]{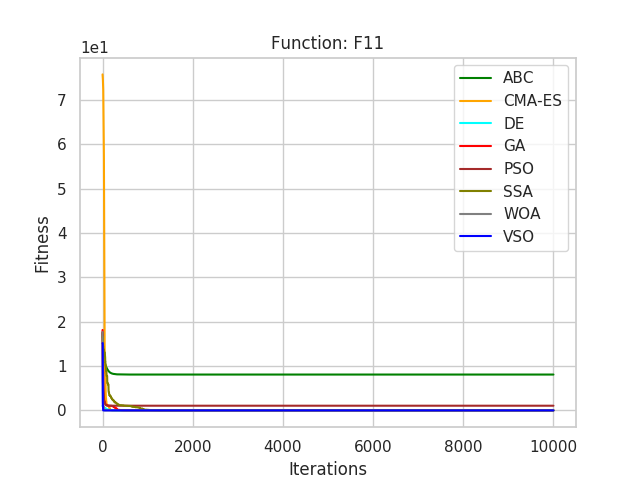}
			
		\end{minipage}%
	}%
	\subfigure[]{
		\begin{minipage}[t]{0.34\linewidth}
			\centering
			\includegraphics[width=2.0in]{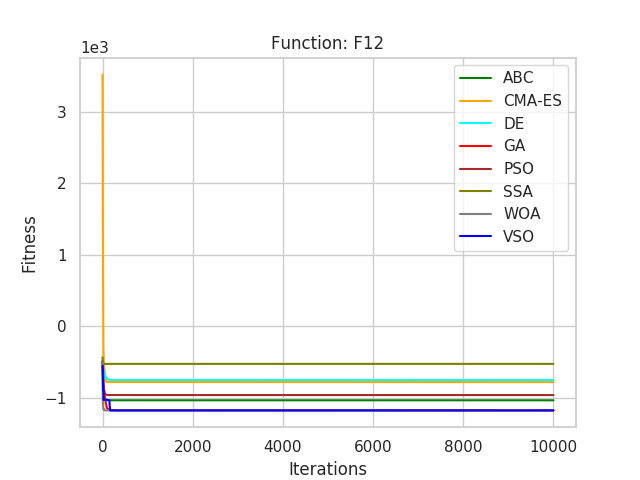}
			
		\end{minipage}
	}%
	
	\subfigure[]{
		\begin{minipage}[t]{0.34\linewidth}
			\centering
			\includegraphics[width=2.0in]{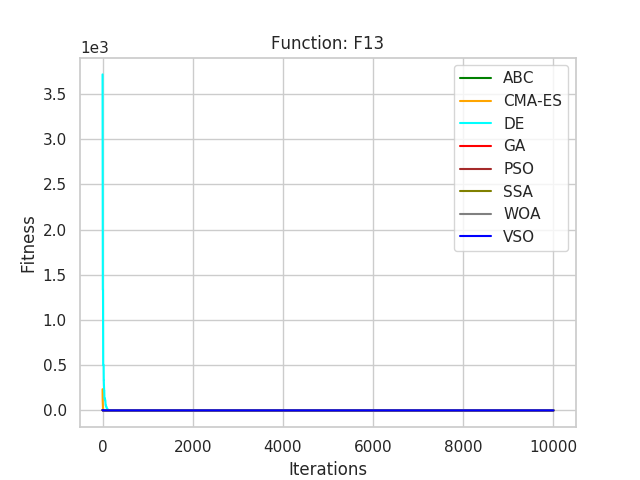}
		\end{minipage}%
	}%
	\subfigure[]{
		\begin{minipage}[t]{0.34\linewidth}
			\centering
			\includegraphics[width=2.0in]{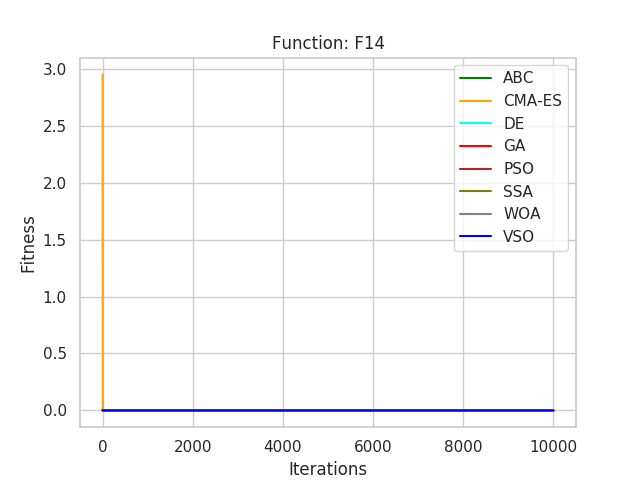}
			
		\end{minipage}%
	}%
	\subfigure[]{
		\begin{minipage}[t]{0.34\linewidth}
			\centering
			\includegraphics[width=2.0in]{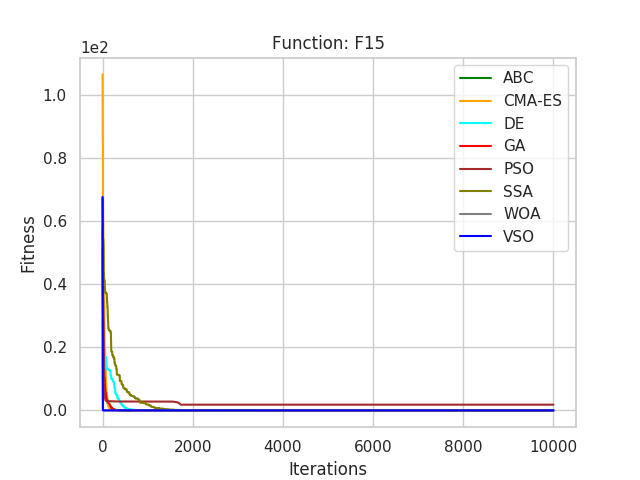}
			
		\end{minipage}
	}%
	\subfigure[]{
		\begin{minipage}[t]{0.34\linewidth}
			\centering
			\includegraphics[width=2.0in]{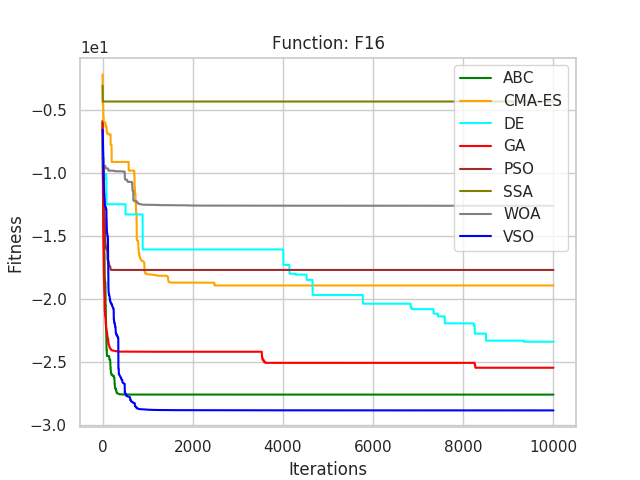}
			
		\end{minipage}
	}%
	\caption{Convergence Test of Results of Classic Benchmarking Functions}
	\label{fig: convergence_plot}
\end{figure*}

~\autoref{fig: convergence_plot} displays the convergence results based on the median fitness of all trials. The results are given as below.

\begin{itemize}
    \item VSO generally converges faster than other algorithms and hence possesses superior convergence capability for such optimization problems;
    \item For the uni-modal functions, it seems that almost all algorithms can quickly converge. This is because most algorithms can achieve small errors as stated in \autoref{tab:classical-func-30}. However, only VSO achieves the exact global optima for these functions which has been discussed in~\Autoref{section: classical_benchmark_functions_test};    
    \item With the more complicated multi-modal functions $F9$-$F16$, it is obvious that the VSO performs very well with respect to the rate of convergence. The convergence rates of some other algorithms decrease, such as GA, DE and SSA on $F9$; DE, ABC and SSA on F10; and CMA-ES and DE on $F16$. WOA has a fast rate of convergence as well. 
\end{itemize}
\subsection{Reliability test}
\begin{figure*}[!htb]
	\centerfloat
	\subfigure[]{
		\begin{minipage}[t]{0.34\linewidth}
			\centering
			\includegraphics[width=2.0in]{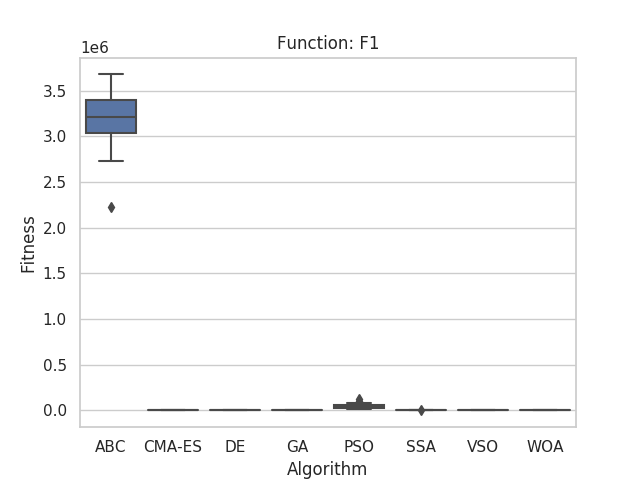}
		\end{minipage}%
	}%
	\subfigure[]{
		\begin{minipage}[t]{0.34\linewidth}
			\centering
			\includegraphics[width=2.0in]{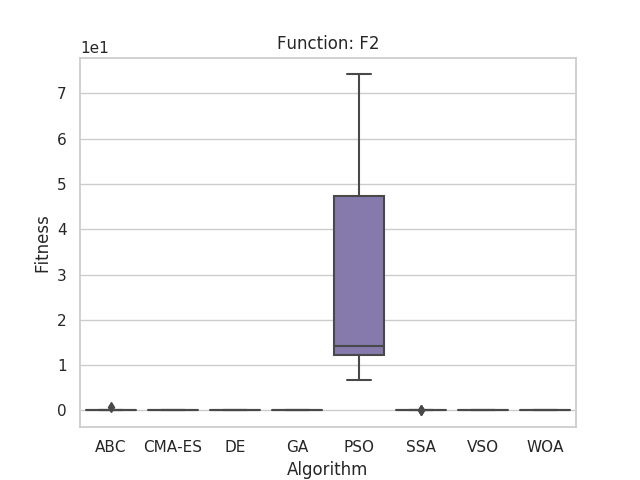}
			
		\end{minipage}%
	}%
	\subfigure[]{
		\begin{minipage}[t]{0.34\linewidth}
			\centering
			\includegraphics[width=2.0in]{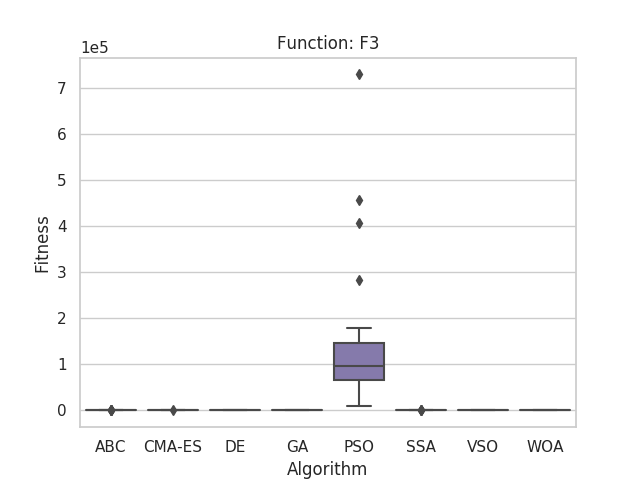}
			
		\end{minipage}
	}%
	\subfigure[]{
		\begin{minipage}[t]{0.34\linewidth}
			\centering
			\includegraphics[width=2.0in]{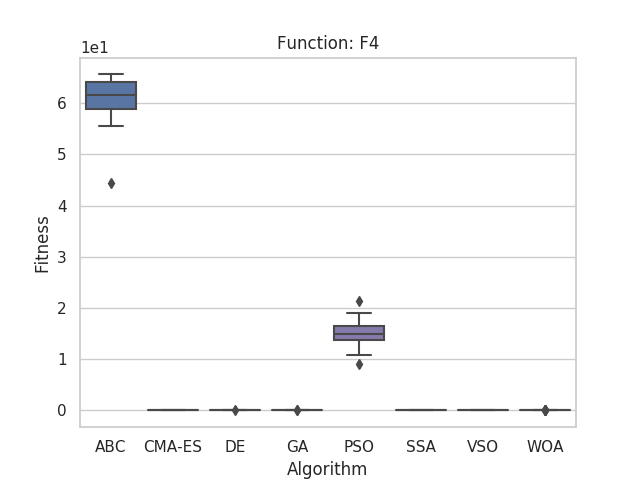}
		\end{minipage}%
	}%
	
	\subfigure[]{
		\begin{minipage}[t]{0.34\linewidth}
			\centering
			\includegraphics[width=2.0in]{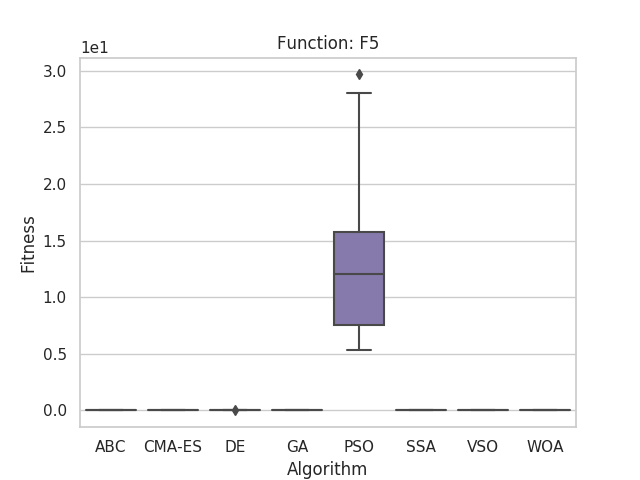}
			
		\end{minipage}%
	}%
	\subfigure[]{
		\begin{minipage}[t]{0.34\linewidth}
			\centering
			\includegraphics[width=2.0in]{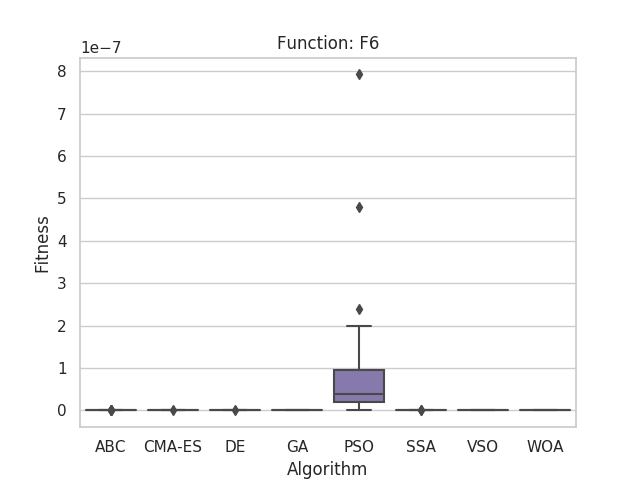}
			
		\end{minipage}
	}%
	\subfigure[]{
		\begin{minipage}[t]{0.34\linewidth}
			\centering
			\includegraphics[width=2.0in]{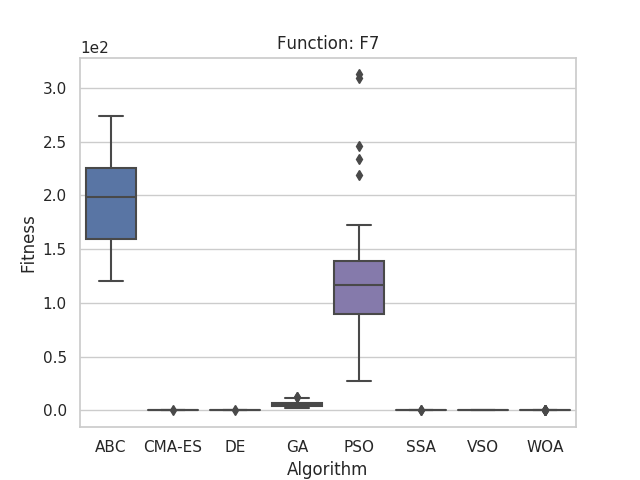}
		\end{minipage}%
	}%
	\subfigure[]{
		\begin{minipage}[t]{0.34\linewidth}
			\centering
			\includegraphics[width=2.0in]{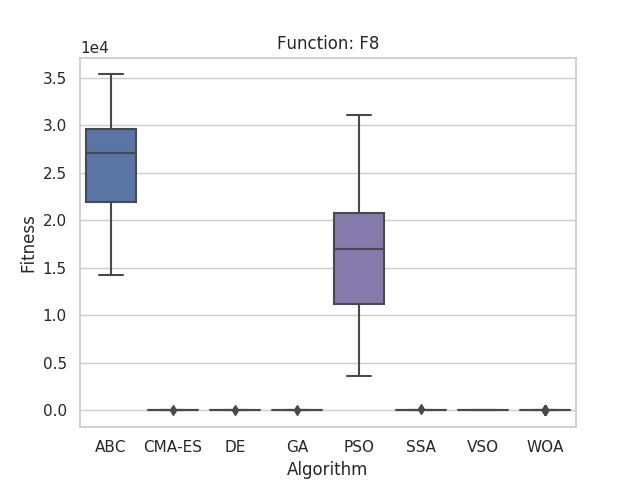}
			
		\end{minipage}%
	}%
	
	\subfigure[]{
		\begin{minipage}[t]{0.34\linewidth}
			\centering
			\includegraphics[width=2.0in]{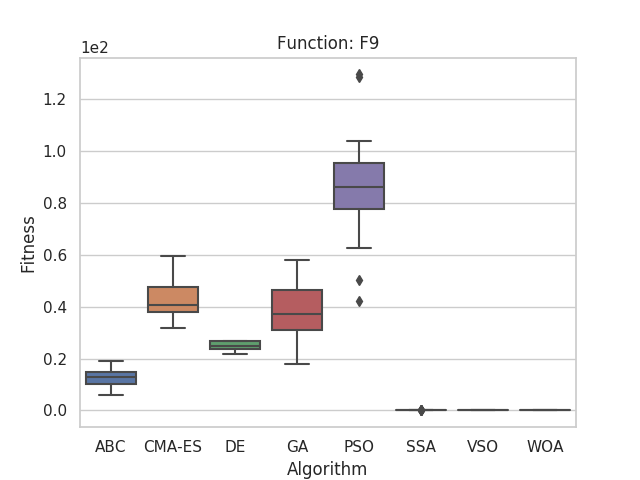}
			
		\end{minipage}
	}%
	\subfigure[]{
		\begin{minipage}[t]{0.34\linewidth}
			\centering
			\includegraphics[width=2.0in]{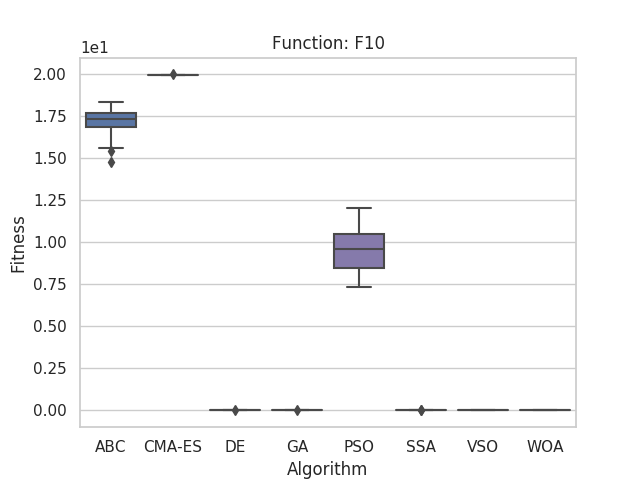}
			
		\end{minipage}
	}%
	\subfigure[]{
		\begin{minipage}[t]{0.34\linewidth}
			\centering
			\includegraphics[width=2.0in]{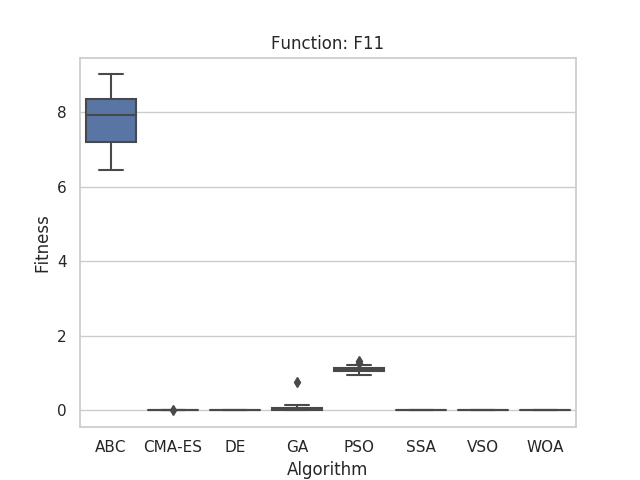}
			
		\end{minipage}
	}%
	\subfigure[]{
		\begin{minipage}[t]{0.34\linewidth}
			\centering
			\includegraphics[width=2.0in]{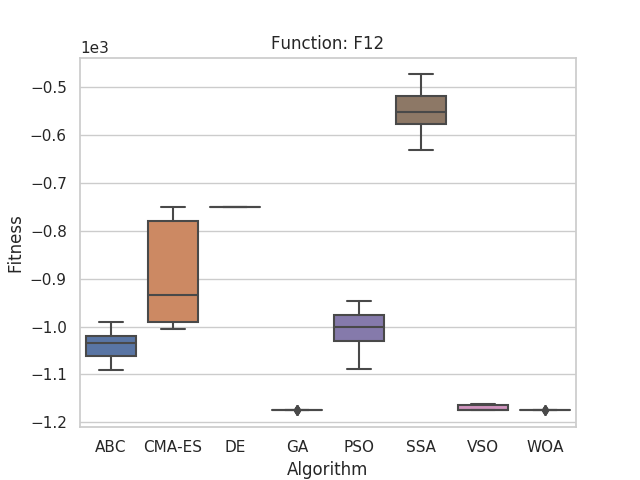}
			
		\end{minipage}
	}%
	
	\subfigure[]{
		\begin{minipage}[t]{0.34\linewidth}
			\centering
			\includegraphics[width=2.0in]{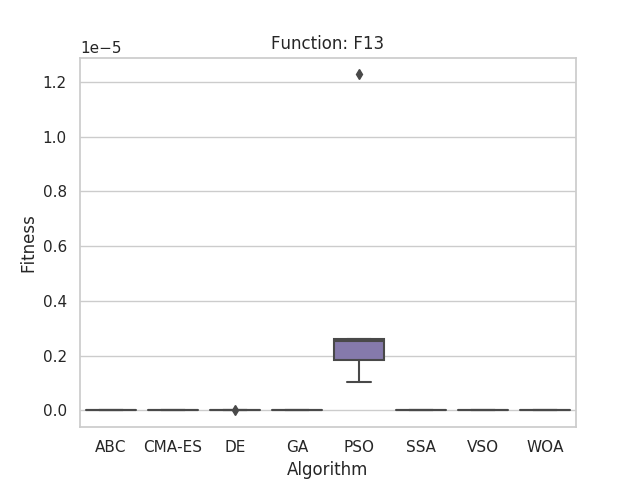}
			
		\end{minipage}
	}%
	\subfigure[]{
		\begin{minipage}[t]{0.34\linewidth}
			\centering
			\includegraphics[width=2.0in]{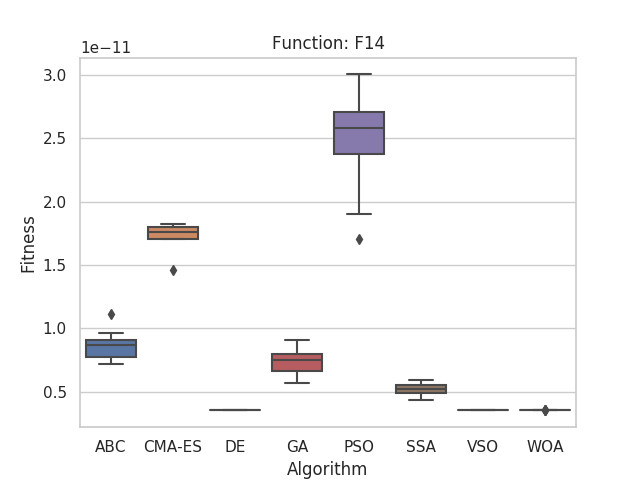}
			
		\end{minipage}
	}%
	\subfigure[]{
		\begin{minipage}[t]{0.34\linewidth}
			\centering
			\includegraphics[width=2.0in]{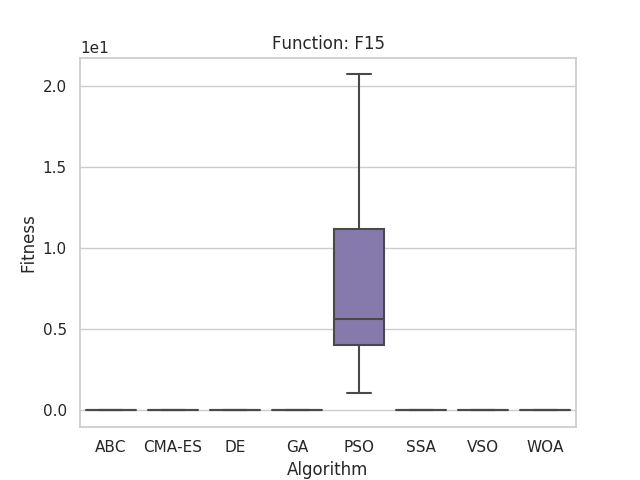}
			
		\end{minipage}
	}%
	\subfigure[]{
		\begin{minipage}[t]{0.34\linewidth}
			\centering
			\includegraphics[width=2.0in]{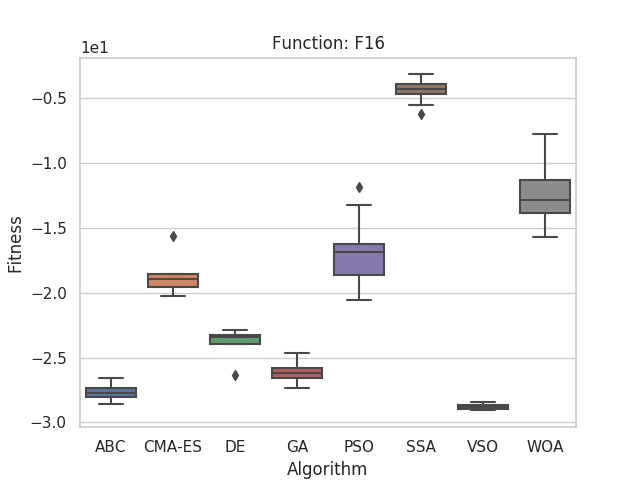}
			
		\end{minipage}
	}%
	\caption{Box Plot of Results of Classic Benchmarking Functions}
	\label{fig: box_plot}
\end{figure*}

~\autoref{fig: box_plot} plots a series of box plots through all runs for the classical benchmark functions for each algorithm. From the obtained results, the following observations can be drawn.

\begin{itemize}
	\item For the uni-modal functions,  the reliability of VSO is impressive over other algorithms. For example, the performances of both ABC and PSO are quite unstable;
	\item For the multi-modal functions, VSO can constantly generate stable results. The only exception is $F12$ where DE achieves the best reliability but it fails to acquire a good solution. The reliability of WOA is followed by VSO. But it becomes much worse in $F16$.

\end{itemize}

\subsection{Scalability Test}\label{section:scalability_test}
In addition to the above low dimensional benchmark functions, a series of evaluations are performed on the medium and high dimensional classical benchmark problems, including $100$, $300$, $500$ and $1000$ dimensions to test the scalability of VSO. To make a thorough comparison, we also employ other algorithms in this evaluation. As aforementioned, the parameter settings of each algorithm are the same as the test on the $30$-$D$ problems. 

From the results listed in \autoref{tab:scalability-test-results} we have the following observations:
\begin{itemize}
	\item VSO achieves the best performance in almost all functions with all dimensions except for $F12$ and $F14$ with $100$ dimensions. In other words, the VSO algorithm ranks first on $59$ out of the total $64$ cases ($\approx$ 92.19\%). More importantly, VSO attains the exact globally optimal solutions for most of the cases. Take $F4$ with $500$ dimensions and $F8$ with $1000$ dimensions as examples. Only VSO obtains the globally optima in both cases. For the latter one, other algorithms except for WOA not only fail to get the exact globally optimal solution but also their values are very bad. Similar cases include $F4$ with $100$ to $1000$ dimensions, $F7$ with $100$ to $1000$ dimensions, $F8$ with $100$ to $1000$ dimensions, etc., in which VSO is the only algorithm with the fitness error as $0$. The findings demonstrate the excellent scalability of our proposed algorithm;
	\item More specifically, VSO shows the advantage of the computational time for some $1000$-$D$ high dimensional problems. For instance, the ranking of VSO for the computational time goes up to the second or even the first place in $F2$, $F3$, $F5$, $F6$, $F12$, $F16$;
	\item As for other algorithms, the solution qualities drop down with increasing dimensions. For instance, for CMA-ES, the mean of fitness of $30$-$D$ $F3$ is $5.25E-284$ as reported in \autoref{tab:classical-func-30}. Nevertheless, the values are $1.74E+01$, $2.24E+04$, $1.14E+05$, and $6.38E+05$ for $100$, $300$, $500$, and $1000$-$D$ problems, respectively. Likewise, the mean value of SSA for $F9$ with $30$-$D$ is $2.25E-13$ while the values become $1.77E+02$, $1.42E+03$, $2.93E+03$ and $6.98E+03$ for $100$, $300$, $500$, and $1000$-$D$ problems, accordingly;
	\item The only competitor is WOA probably due to its sophisticated design of the searching strategies as inspired by searching for prey and attacking the prey of the whales~\cite{RN18}.

\end{itemize}

\autoref{tab:summary-scalability-test} summaries the results of the scalability test.
\clearpage
\begingroup
\setlength{\LTleft}{-20cm plus -1fill}
\setlength{\LTright}{\LTleft}

{\footnotesize
\begin{longtable}[!htb]{@{}cclrrrrrrrr@{}}
	\caption{Results of Scalability Test}
	\label{tab:scalability-test-results}\\
	\toprule
	Function &
	Dimension &
	Metric &
	ABC &
	CMA-ES &
	DE &
	GA &
	PSO &
	SSA &
	VSO &
	WOA \\* \midrule
	\endhead
	\bottomrule
	\endfoot
	\endlastfoot
	F1 &
	100 &
	Mean &
	1.99E+07 &
	2.17E-145 &
	2.85E-05 &
	7.78E+01 &
	7.12E+05 &
	8.46E-21 &
	0.00E+00 &
	0.00E+00 \\
	&
	&
	Time(s) &
	12.00 &
	1284.26 &
	208.53 &
	98.58 &
	6.51 &
	11.59 &
	22.12 &
	73.90 \\
	&
	&
	Fitness Rank &
	8 &
	3 &
	5 &
	6 &
	7 &
	4 &
	\cellcolor[HTML]{E7E6E6}\textbf{1} &
	\cellcolor[HTML]{E7E6E6}\textbf{1} \\
	&
	&
	Time Rank &
	3 &
	8 &
	7 &
	6 &
	\cellcolor[HTML]{E7E6E6}\textbf{1} &
	2 &
	4 &
	5 \\
	&
	300 &
	Mean &
	7.61E+07 &
	1.59E-61 &
	5.41E+00 &
	6.08E+05 &
	3.09E+06 &
	1.15E-05 &
	0.00E+00 &
	0.00E+00 \\
	&
	&
	Time(s) &
	8.02 &
	1767.10 &
	678.84 &
	240.96 &
	10.88 &
	13.55 &
	35.24 &
	138.95 \\
	&
	&
	Fitness Rank &
	8 &
	3 &
	5 &
	6 &
	7 &
	4 &
	\cellcolor[HTML]{E7E6E6}\textbf{1} &
	\cellcolor[HTML]{E7E6E6}\textbf{1} \\
	&
	&
	Time Rank &
	\cellcolor[HTML]{E7E6E6}\textbf{1} &
	8 &
	7 &
	6 &
	2 &
	3 &
	4 &
	5 \\
\bottomrule
\end{longtable}
}

\endgroup
\begin{table}[!htb]
	\footnotesize
	\centering
	\caption{Summary of Scalability Test}
	\label{tab:summary-scalability-test}
    %
 \\ \midrule
	ABC    & 6.84 & 3.88 & 8                                  & 3                                  \\
	CMA-ES & 4.42 & 7.91 & 3                                  & 8                                  \\
	DE     & 5.27 & 6.41 & 6                                  & 7                                  \\
	GA     & 4.72 & 4.92 & 4                                  & 6                                  \\
	PSO    & 6.36 & 2.06 & 7                                  & \cellcolor[HTML]{E7E6E6}\textbf{1} \\
	SSA    & 4.81 & 2.88 & 5                                  & 2                                  \\
	VSO    & 1.13 & 3.89 & \cellcolor[HTML]{E7E6E6}\textbf{1} & 4                                  \\
	WOA    & 1.69 & 4.06 & 2                                  & 5                                  \\ \bottomrule
    \end{tabular}
\end{table}

\subsection{Evaluation on VSO without Imported Infection}\label{section:vso_no_im}

To investigate the performance of VSO without the imported infection operation, which is an additional function, another evaluation was conducted on the same set of classical and CEC benchmark functions. From \autoref{tab:vso-without-im-results} \& \ref{tab:vso-without-im-scalability}, we can observe that:
\begin{table}[!htb]
	\footnotesize
	\centering
	\caption{Results of Benchmarking Functions by VSO without Imported Infection}
	\label{tab:vso-without-im-results}
	\begin{tabular}{@{}lcrrrrr@{}}
		\toprule
		Function & Dimension & \multicolumn{1}{c}{Mean} & \multicolumn{1}{c}{Std} & \multicolumn{1}{c}{Best} & \multicolumn{1}{c}{Worst} & \multicolumn{1}{c}{Time(s)} \\ \midrule
		F1    & 30 & 0.00E+00  & 0.00E+00 & 0.00E+00  & 0.00E+00  & 10.15  \\
		F2    & 30 & 0.00E+00  & 0.00E+00 & 0.00E+00  & 0.00E+00  & 75.92  \\
		F3    & 30 & 0.00E+00  & 0.00E+00 & 0.00E+00  & 0.00E+00  & 52.46  \\
		F4    & 30 & 0.00E+00  & 0.00E+00 & 0.00E+00  & 0.00E+00  & 10.13  \\
		F5    & 30 & 0.00E+00  & 0.00E+00 & 0.00E+00  & 0.00E+00  & 35.65  \\
		F6    & 30 & 0.00E+00  & 0.00E+00 & 0.00E+00  & 0.00E+00  & 44.40  \\
		F7    & 30 & 0.00E+00  & 0.00E+00 & 0.00E+00  & 0.00E+00  & 40.06  \\
		F8    & 30 & 0.00E+00  & 0.00E+00 & 0.00E+00  & 0.00E+00  & 128.28 \\
		F9    & 30 & 0.00E+00  & 0.00E+00 & 0.00E+00  & 0.00E+00  & 11.86  \\
		F10   & 30 & 4.44E-16  & 0.00E+00 & 4.44E-16  & 4.44E-16  & 14.87  \\
		F11   & 30 & 0.00E+00  & 0.00E+00 & 0.00E+00  & 0.00E+00  & 14.94  \\
		F12   & 30 & -1.00E+03 & 2.38E+01 & -1.05E+03 & -9.63E+02 & 63.79  \\
		F13   & 30 & 0.00E+00  & 0.00E+00 & 0.00E+00  & 0.00E+00  & 13.75  \\
		F14   & 30 & 3.51E-12  & 4.11E-26 & 3.51E-12  & 3.51E-12  & 58.31  \\
		F15   & 30 & 0.00E+00  & 0.00E+00 & 0.00E+00  & 0.00E+00  & 42.11  \\
		F16   & 30 & -1.93E+01 & 2.73E+00 & -2.34E+01 & -1.48E+01 & 86.54  \\
		CEC1  & 30 & 1.92E+06  & 4.75E+05 & 1.13E+06  & 2.67E+06  & 11.39  \\
		CEC2  & 30 & 1.30E+04  & 1.42E+04 & 1.45E+03  & 3.45E+04  & 10.89  \\
		CEC3  & 30 & 5.25E-02  & 3.27E-02 & 9.03E-03  & 1.12E-01  & 10.93  \\
		CEC4  & 30 & 8.01E+01  & 3.71E+01 & 4.99E+00  & 1.42E+02  & 12.02  \\
		CEC5  & 30 & 2.00E+01  & 7.14E-02 & 2.00E+01  & 2.02E+01  & 11.74  \\
		CEC6  & 30 & 3.05E+01  & 4.08E+00 & 1.95E+01  & 3.45E+01  & 45.94  \\
		CEC7  & 30 & 1.52E-02  & 1.76E-02 & 1.02E-12  & 4.92E-02  & 11.85  \\
		CEC8  & 30 & 1.33E+02  & 2.09E+01 & 1.00E+02  & 1.65E+02  & 11.21  \\
		CEC9  & 30 & 1.85E+02  & 1.82E+01 & 1.64E+02  & 2.23E+02  & 11.77  \\
		CEC10 & 30 & 3.38E+03  & 7.10E+02 & 2.40E+03  & 4.44E+03  & 11.04  \\
		CEC11 & 30 & 4.41E+03  & 5.31E+02 & 3.85E+03  & 5.42E+03  & 11.50  \\
		CEC12 & 30 & 1.05E+00  & 3.46E-01 & 4.64E-01  & 1.67E+00  & 20.79  \\
		CEC13 & 30 & 5.18E-01  & 1.28E-01 & 2.80E-01  & 7.47E-01  & 10.51  \\
		CEC14 & 30 & 2.88E-01  & 4.45E-02 & 2.38E-01  & 3.84E-01  & 10.51  \\
		CEC15 & 30 & 3.22E+01  & 9.39E+00 & 1.70E+01  & 5.08E+01  & 11.10  \\
		CEC16 & 30 & 1.22E+01  & 4.07E-01 & 1.14E+01  & 1.29E+01  & 11.16  \\
		CEC17 & 30 & 9.86E+04  & 6.53E+04 & 1.58E+04  & 2.45E+05  & 12.53  \\
		CEC18 & 30 & 8.46E+03  & 1.06E+04 & 3.21E+02  & 2.62E+04  & 12.00  \\
		CEC19 & 30 & 1.68E+01  & 1.34E+00 & 1.37E+01  & 1.91E+01  & 18.34  \\
		CEC20 & 30 & 3.48E+02  & 7.91E+01 & 1.92E+02  & 4.86E+02  & 11.96  \\
		CEC21 & 30 & 4.98E+04  & 3.32E+04 & 7.41E+03  & 1.06E+05  & 11.40  \\
		CEC22 & 30 & 6.05E+02  & 1.20E+02 & 4.38E+02  & 7.86E+02  & 12.42  \\
		CEC23 & 30 & 2.00E+02  & 0.00E+00 & 2.00E+02  & 2.00E+02  & 16.59  \\
		CEC24 & 30 & 2.00E+02  & 0.00E+00 & 2.00E+02  & 2.00E+02  & 14.84  \\
		CEC25 & 30 & 2.00E+02  & 0.00E+00 & 2.00E+02  & 2.00E+02  & 15.40  \\
		CEC26 & 30 & 1.00E+02  & 1.64E-01 & 1.00E+02  & 1.01E+02  & 57.77  \\
		CEC27 & 30 & 2.00E+02  & 0.00E+00 & 2.00E+02  & 2.00E+02  & 58.32  \\
		CEC28 & 30 & 2.00E+02  & 0.00E+00 & 2.00E+02  & 2.00E+02  & 19.64  \\
		CEC29 & 30 & 2.00E+02  & 0.00E+00 & 2.00E+02  & 2.00E+02  & 22.62  \\
		CEC30 & 30 & 2.00E+02  & 0.00E+00 & 2.00E+02  & 2.00E+02  & 16.57  \\ \bottomrule
	\end{tabular}
\end{table}
\begin{itemize}
	\item Through a comparison of \autoref{tab:classical-func-30} and \ref{tab:vso-without-im-results}, the mean values of fitness by VSO without any imported infection operation are same as VSO with DE except for $F12$ and $F16$;
	\item As for the complicated CEC benchmark functions, the performances of two approaches are same for $5$ cases, i.e. $CEC23$, $CEC24$, $CEC25$, $CEC27$, and $CEC28$. VSO without DE even achieves better in the cases of $CEC1$, $CEC17$, $CEC29$, and $CEC30$. For the remaining $21$ functions, the VSO algorithm with the imported infection powered by DE is readily better;
	\item In terms of the computational time, the average time of running all $46$ classical and CEC $30$-$D$ functions for VSO with and without DE are $50.84s$ and $26.69s$, respectively. This means the introduction of such imported infection operation almost doubles the computational time;
	\item Taking $100$-$D$ classical benchmark functions as examples, the scalability of VSO without the imported infection mechanism is also impressive.

\end{itemize}

On the other hand, a more thorough investigation should be conducted in the future work on what specific condition(s) and how this additional operator can actually help to enhance the search performance of VSO in handling various complex real-world applications. Furthermore, other meta-heuristic algorithms can be studied as the algorithm in the imported infection operation. 

\begin{table}[!htb]
\footnotesize
\centering
\caption{Results of Scalability Test by VSO without Imported Infection}
\label{tab:vso-without-im-scalability}
\begin{tabular}{@{}lcrrrrr@{}}
\toprule
Function & Dimension & \multicolumn{1}{c}{Mean} & \multicolumn{1}{c}{Std} & \multicolumn{1}{c}{Best} & \multicolumn{1}{c}{Worst} & \multicolumn{1}{c}{Time(s)} \\ \midrule
F01 & 100 & 0.00E+00  & 0.00E+00 & 0.00E+00  & 0.00E+00  & 13.04  \\
F02 & 100 & 0.00E+00  & 0.00E+00 & 0.00E+00  & 0.00E+00  & 240.49 \\
F03 & 100 & 0.00E+00  & 0.00E+00 & 0.00E+00  & 0.00E+00  & 157.39 \\
F04 & 100 & 0.00E+00  & 0.00E+00 & 0.00E+00  & 0.00E+00  & 12.95  \\
F05 & 100 & 0.00E+00  & 0.00E+00 & 0.00E+00  & 0.00E+00  & 101.21 \\
F06 & 100 & 0.00E+00  & 0.00E+00 & 0.00E+00  & 0.00E+00  & 129.01 \\
F07 & 100 & 0.00E+00  & 0.00E+00 & 0.00E+00  & 0.00E+00  & 111.41 \\
F08 & 100 & 0.00E+00  & 0.00E+00 & 0.00E+00  & 0.00E+00  & 634.53 \\
F09 & 100 & 0.00E+00  & 0.00E+00 & 0.00E+00  & 0.00E+00  & 15.03  \\
F10 & 100 & 4.44E-16  & 0.00E+00 & 4.44E-16  & 4.44E-16  & 18.14  \\
F11 & 100 & 0.00E+00  & 0.00E+00 & 0.00E+00  & 0.00E+00  & 21.25  \\
F12 & 100 & -3.28E+03 & 4.98E+01 & -3.39E+03 & -3.22E+03 & 196.11 \\
F13 & 100 & 0.00E+00  & 0.00E+00 & 0.00E+00  & 0.00E+00  & 20.28  \\
F14 & 100 & 4.66E-42  & 1.41E-54 & 4.66E-42  & 4.66E-42  & 176.46 \\
F15 & 100 & 0.00E+00  & 0.00E+00 & 0.00E+00  & 0.00E+00  & 119.50 \\
F16 & 100 & -4.71E+01 & 5.59E+00 & -5.96E+01 & -4.09E+01 & 269.93 \\ \bottomrule
\end{tabular}
\end{table}

\section{Real-world Application I: Financial Portfolio Optimization}\label{section:portfolio_optimization}
\subsection{Problem Description}
Portfolio optimization is one of the most important problems in finance. Investors usually want to maximize returns and minimize risks through allocating a fixed amount of capital into a collection of assets.

According to the mean-variance model, which is a well-known and widely-used portfolio optimization theory formulated by Markowitz~\cite{RN51}, the variance is a risk measure. The optimization problem is presented in~\eqref{eq:eq14}-\eqref{eq:eq15} as below.

\begin{equation}
	max\ E(R(x))=\sum_{i=1}^nx_i u_i
	\label{eq:eq14}
\end{equation}
\begin{equation}
	min\ V(R(x))=\sum_{i=1}^n\sum_{j=1}^nx_i x_j\sigma_{ij}
	\label{eq:eq15}
\end{equation}
\begin{equation}\nonumber
	subject  \; to:\, x \in X = \{x \in R \,| \sum_{i=1}^nx_i = 1, x_i \geq 0\}
\end{equation}
where $x_i$ is the proportion weight of the initial capital that will be allocated in the $i^{th}$ asset, $u_i$ is the return of the $i^{th}$ asset, $\sigma_{ij}$ stand for the covariance of returns of the $i^{th}$ and $j^{th}$ assets, $E(R(x))$ and $V(R(x))$ are the expected return and variance of the whole portfolio, respectively.
	
To optimize above two objectives simultaneously, we combine them into one single objective function as shown in~\eqref{eq:eq16}.

\begin{equation}
	max\ SR = \frac{E(R(x)) - R_f}{V(R(x))}
	\label{eq:eq16}
\end{equation}
where SR is called the sharpe ratio that represents the return and risk simultaneously of the portfolio, $R_f$ is a risk-free rate.

Also, the sharpe ratio has been one of the most important measurement tools to evaluate the performance of investment portfolio in the real-world financial industry.

Since the VSO and other comparative algorithms are designated for solving minimization problems, the problem should be changed to the minimization problem as given in~\eqref{eq:eq17}.

\begin{equation}
min\ fitness = \frac{1}{SR}
\label{eq:eq17}
\end{equation}

\begin{equation}\nonumber
subject  \; to:\, SR=10^{-10}  \; if  \; SR \leq 0
\end{equation}

As it is very possible that the return is zero or even negative in the financial market, a very small number $10^{-10}$ is assigned to the SR for this case.

In order to avoid handling the equality constraint, the solution can be converted to the unconstrained form as shown in \ref{eq:eq18}.

\begin{equation}
x_i' = \frac{x_i}{\sum_{i=1}^n|x_i|}
\label{eq:eq18}
\end{equation}

\subsection{Experimental Setting}
Considering that the U.S. stock market is the biggest developed market and the Chinese stock market is the biggest emerging market all over the world, we select these two markets as our experimental targets. For the U.S. market, $S\&P500$ represents $500$ large companies listed on stock exchanges in the U.S. Likewise, $CSI300$ constituent stocks are the top $300$ stocks traded on the Shanghai Stock Exchange and the Shenzhen Stock Exchange. As the lists of both $S\&P500$ and $CSI300$ were adjusted from time to time, we selected the maximum number up to $250$ stocks in each group according to the order of their stock symbols to make a fair comparison. The full stocks list is illustrated in \autoref{tab:stocks_list}.
	
Following the previous practice, the information of mean and covariance is acquired from the historical data. In the experiment, we calculated such values through the $5$-year historical daily data of the candidate stocks, i.e. from 1 Jan 2015 to 31 Dec 2019 excluding non-transaction days. More specifically, the average daily return on the historical data is computed as the expected return of each stock.  

We also tried different number of stocks, i.e. $10$, $30$, $100$ and $250$, to further investigate the scalability of the algorithm on this practical application. More specially, in addition to longing the stocks, Additionally, we studied a real-world scenario in which short-selling is allowed, i.e. $x_i$ can be negative, which enlarges the searching space of the problem.  The US $5$-$year$ treasury yield of $2.57\%$ and China $3$-$year$ fixed deposit interest rate of $4.22\%$ are performed as the risk-free rates for $S\&P500$ and $CSI300$, accordingly.

We utilized the same parameters set for each algorithm as the benchmark functions test. But we set the maximum iteration as $10^3$ due to a large amount of the data.

\subsection{Results and Discussion}
Different from the benchmark functions tests, the portfolio optimization is a maximization problem here. From \autoref{tab:portfolio-results}-\ref{tab:porfolio-summary}, we can see that:
\begin{itemize}
	\item For the group of $CSI300$, VSO achieves best in $6$ cases. In particular, VSO gets the impressive sharpe ratio of $2.802$ in the case of $CSI300$ Long/Short $250$ stocks, which totally beat other algorithms (the second best $1.7184$ is generated by ABC). In this group, the performance of SSA is also good. Regarding the case of Long $250$ stocks, SSA obtains $1.8787$ v.s. $1.6666$ by VSO;
	\item For the group of $S\&P500$, VSO performs best in $5$ cases. Similarly, VSO acquires the result, i.e. $3.6203$, that is much better than others for the case of Long/Short $250$ stocks;
	\item Above phenomenon may imply that VSO is good at optimizing high-dimensional problems with large searching spaces;
\end{itemize}
\begingroup
\setlength{\LTleft}{-20cm plus -1fill}
\setlength{\LTright}{\LTleft}

{\footnotesize

\begin{longtable}[!htb]{@{}llrrrrrrrr@{}}
	\caption{Results of Financial Portfolio Optimization}
	\label{tab:portfolio-results}\\
	\toprule
	{\color[HTML]{000000} Problem} &
	{\color[HTML]{000000} Metric} &
	\multicolumn{1}{c}{{\color[HTML]{000000} ABC}} &
	\multicolumn{1}{c}{{\color[HTML]{000000} CMA-ES}} &
	\multicolumn{1}{c}{{\color[HTML]{000000} DE}} &
	\multicolumn{1}{c}{{\color[HTML]{000000} GA}} &
	\multicolumn{1}{c}{{\color[HTML]{000000} PSO}} &
	\multicolumn{1}{c}{{\color[HTML]{000000} SSA}} &
	\multicolumn{1}{c}{{\color[HTML]{000000} VSO}} &
	\multicolumn{1}{c}{{\color[HTML]{000000} WOA}} \\* \midrule
	\endhead
	\bottomrule
	\endfoot
	\endlastfoot
	{\color[HTML]{000000} CSI 300} &
	{\color[HTML]{000000} Mean} &
	{\color[HTML]{000000} 1.224294} &
	{\color[HTML]{000000} 1.154355} &
	{\color[HTML]{000000} 1.224411} &
	{\color[HTML]{000000} 1.223784} &
	{\color[HTML]{000000} 1.181852} &
	{\color[HTML]{000000} 1.224364} &
	{\color[HTML]{000000} 1.224416} &
	{\color[HTML]{000000} 1.182489} \\
	{\color[HTML]{000000} Long 10 stocks} &
	{\color[HTML]{000000} Std} &
	{\color[HTML]{000000} 0.000042} &
	{\color[HTML]{000000} 0.017363} &
	{\color[HTML]{000000} 0.000005} &
	{\color[HTML]{000000} 0.000555} &
	{\color[HTML]{000000} 0.021924} &
	{\color[HTML]{000000} 0.000017} &
	{\color[HTML]{000000} 0.000005} &
	{\color[HTML]{000000} 0.067856} \\
	{\color[HTML]{000000} } &
	{\color[HTML]{000000} Best} &
	{\color[HTML]{000000} 1.224359} &
	{\color[HTML]{000000} 1.170395} &
	{\color[HTML]{000000} 1.224418} &
	{\color[HTML]{000000} 1.224194} &
	{\color[HTML]{000000} 1.214714} &
	{\color[HTML]{000000} 1.224398} &
	{\color[HTML]{000000} 1.224423} &
	{\color[HTML]{000000} 1.224423} \\
	{\color[HTML]{000000} } &
	{\color[HTML]{000000} Worst} &
	{\color[HTML]{000000} 1.224248} &
	{\color[HTML]{000000} 1.123408} &
	{\color[HTML]{000000} 1.224403} &
	{\color[HTML]{000000} 1.222685} &
	{\color[HTML]{000000} 1.149505} &
	{\color[HTML]{000000} 1.224348} &
	{\color[HTML]{000000} 1.224409} &
	{\color[HTML]{000000} 1.047452} \\
	{\color[HTML]{000000} } &
	{\color[HTML]{000000} Time(s)} &
	{\color[HTML]{000000} 7.73} &
	{\color[HTML]{000000} 3.92} &
	{\color[HTML]{000000} 7.14} &
	{\color[HTML]{000000} 6.44} &
	{\color[HTML]{000000} 5.06} &
	{\color[HTML]{000000} 5.32} &
	{\color[HTML]{000000} 7.90} &
	{\color[HTML]{000000} 6.66} \\
\bottomrule
\end{longtable}
}

\endgroup

\begin{table}[tb]
	\footnotesize
	\centering
	\caption{Summary of Evaluations for Financial Portfolio Optimization}
	\label{tab:porfolio-summary}
	%
} \\ \midrule
		{\color[HTML]{000000} ABC} &
		{\color[HTML]{000000} 3.94} &
		{\color[HTML]{000000} 4.06} &
		{\color[HTML]{000000} 3} &
		{\color[HTML]{000000} 3} \\
		{\color[HTML]{000000} CMA-ES} &
		{\color[HTML]{000000} 4.63} &
		{\color[HTML]{000000} 6.56} &
		{\color[HTML]{000000} 5} &
		{\color[HTML]{000000} 7} \\
		{\color[HTML]{000000} DE} &
		{\color[HTML]{000000} 5.75} &
		{\color[HTML]{000000} 6.94} &
		{\color[HTML]{000000} 6} &
		{\color[HTML]{000000} 8} \\
		{\color[HTML]{000000} GA} &
		{\color[HTML]{000000} 4.13} &
		{\color[HTML]{000000} 5.31} &
		{\color[HTML]{000000} 4} &
		{\color[HTML]{000000} 6} \\
		{\color[HTML]{000000} PSO} &
		{\color[HTML]{000000} 7.06} &
		{\color[HTML]{000000} 1.44} &
		{\color[HTML]{000000} 8} &
		\cellcolor[HTML]{E7E6E6}{\color[HTML]{000000} \textbf{1}} \\
		{\color[HTML]{000000} SSA} &
		{\color[HTML]{000000} 2.56} &
		{\color[HTML]{000000} 2.50} &
		{\color[HTML]{000000} 2} &
		{\color[HTML]{000000} 2} \\
		{\color[HTML]{000000} VSO} &
		{\color[HTML]{000000} 1.38} &
		{\color[HTML]{000000} 4.38} &
		\cellcolor[HTML]{E7E6E6}{\color[HTML]{000000} \textbf{1}} &
		{\color[HTML]{000000} 4} \\
		{\color[HTML]{000000} WOA} &
		{\color[HTML]{000000} 6.56} &
		{\color[HTML]{000000} 4.81} &
		{\color[HTML]{000000} 7} &
		{\color[HTML]{000000} 5} \\ \bottomrule
	\end{tabular}
\end{table}

\section{Real-world Application II: Optimization of Hyper-parameters of Support Vector Machines}\label{section:SVM}
\subsection{Problem Description}
SVMs are widely adopted machine learning algorithms particularly useful for some limited sample datasets within the framework of the statistical learning theory. According to the literature, SVMs have achieved impressive success in various applications, such as image classification~\cite{RN52}, natural language processing~\cite{RN53}, and financial prediction~\cite{RN54}, etc.

In practice, the performance of SVMs usually depends on its hyper-parameters. There are two major types of algorithms in SVMs: classification and regression. In this experiment, we apply SVMs to classify some real-world practical datasets. 

The mathematical expressions of SVM is shown as in~\eqref{eq:eq19}.
\begin{eqnarray}
	\begin{split}
		\max_\alpha \sum_j \alpha_j - \frac{1}{2} \sum_{j,k} \alpha_j, \alpha_k y_j y_k k(x_j, x_k)\\ 
         subject \; to: 0 \leq \alpha_j \leq C \; and \; \sum_{j} \alpha_j y_j = 0    
	\end {split}
	\label{eq:eq19}
\end{eqnarray}
where $C$ is the tunable penalty factor and $K$ is the kernal function. 
Due to the outstanding performance of RBF kernel function, it is used in this test as stated in~\eqref{eq:eq20}.
\begin{eqnarray}
	k(x_j, x_k) = exp\Bigg( - \frac{ \parallel x_j - x_k \parallel ^2 }{2 \sigma^2} \Bigg)
	\label{eq:eq20}
\end{eqnarray}
where $\sigma$ is another tunable parameter. 
Using this kernel in the SVM classifier, we can get the decision function as shown in~\eqref{eq:eq21}.
\begin{eqnarray}
	f(x) = \textrm{sign}\Bigg[ \sum_i \alpha_i y_i
    	exp\Bigg( - \frac{ \parallel x - x_i \parallel ^2 }{2 \sigma^2} \Bigg)
    	+ b \Bigg] 
    	\label{eq:eq21}
\end{eqnarray}

In this test, we have to optimize two hyper-parameters: the penalty factor $C$ and $\sigma$ for classification problems.
\subsection{Experimental Setting}
There are five datasets involved:
\begin{itemize}
	\item Australian Credit Approval: A well-known dataset that concerns credit card applications approval in Australia~\cite{RN49};
	\item HCC Survival: HCC dataset was obtained at a University Hospital in Portugal and contais several demographic, risk factors, laboratory and overall survival features of 165 real patients diagnosed with HCC~\cite{RN47};
	\item Iris:This is perhaps the best known dataset to be found in machine learning. It is to classify  type of iris plant~\cite{RN49};
	\item Somerville Happiness Survey: A dataset about life survey~\cite{RN48}.
	\item Wine: This dataset is the results of a chemical analysis of wines grown in the same region in Italy but derived from three different cultivars~\cite{RN49};
\end{itemize}

All datasets as listed above are publicly available at~\cite{RN49}. As for the searching space, we set $C, \sigma \in [10^{-5},10^5]$. 
The accuracy of $10$-$fold$ cross validation is computed as the fitness in the evaluation. The maximum iteration is set as $500$.
\subsection{Results and Discussion}
The detailed results are illustrated into \autoref{tab:svm-results} where the mean fitness represents the average classification accuracy of 30 runs. The findings are stated as follows.
\begin{itemize}
	\item VSO outperforms over all other algorithms on $4$ out of $5$ datasets. For example, VSO achieves $83.83\%$ of mean accuracy for the first dataset;
    \item The performance of CMA-ES is very bad in this test. On the other hand, ABC that performs unwell in previous benchmark functions tests becomes not bad here;
    \item Although VSO can get an enhancement of accuracy, it does not show a big advantage compared with other candidates in this low-dimensional problem optimization.
\end{itemize}
\begingroup
\setlength{\LTleft}{-20cm plus -1fill}
\setlength{\LTright}{\LTleft}

{\footnotesize
\begin{longtable}[!htb]{llrrrrrrrr}
	\caption{Results of SVMs Optimization}
	\label{tab:svm-results}\\
	\hline
	{\color[HTML]{000000} Dataset} &
	{\color[HTML]{000000} Metric} &
	\multicolumn{1}{c}{{\color[HTML]{000000} ABC}} &
	\multicolumn{1}{c}{{\color[HTML]{000000} CMA-ES}} &
	\multicolumn{1}{c}{{\color[HTML]{000000} DE}} &
	\multicolumn{1}{c}{{\color[HTML]{000000} GA}} &
	\multicolumn{1}{c}{{\color[HTML]{000000} PSO}} &
	\multicolumn{1}{c}{{\color[HTML]{000000} SSA}} &
	\multicolumn{1}{c}{{\color[HTML]{000000} VSO}} &
	\multicolumn{1}{c}{{\color[HTML]{000000} WOA}} \\ \hline
	\endhead
	\hline
	\endfoot
	\endlastfoot
	{\color[HTML]{000000} Australian Credit} &
	{\color[HTML]{000000} Mean} &
	{\color[HTML]{000000} 0.8023} &
	{\color[HTML]{000000} 0.6058} &
	{\color[HTML]{000000} 0.7203} &
	{\color[HTML]{000000} 0.7119} &
	{\color[HTML]{000000} 0.7470} &
	{\color[HTML]{000000} 0.8075} &
	{\color[HTML]{000000} 0.8383} &
	{\color[HTML]{000000} 0.8203} \\
	{\color[HTML]{000000} Approval} &
	{\color[HTML]{000000} Std} &
	{\color[HTML]{000000} 0.0159} &
	{\color[HTML]{000000} 0.0000} &
	{\color[HTML]{000000} 0.0000} &
	{\color[HTML]{000000} 0.0550} &
	{\color[HTML]{000000} 0.0628} &
	{\color[HTML]{000000} 0.0521} &
	{\color[HTML]{000000} 0.0020} &
	{\color[HTML]{000000} 0.0062} \\
	{\color[HTML]{000000} } &
	{\color[HTML]{000000} Best} &
	{\color[HTML]{000000} 0.8203} &
	{\color[HTML]{000000} 0.6058} &
	{\color[HTML]{000000} 0.7203} &
	{\color[HTML]{000000} 0.8217} &
	{\color[HTML]{000000} 0.8246} &
	{\color[HTML]{000000} 0.8435} &
	{\color[HTML]{000000} 0.8406} &
	{\color[HTML]{000000} 0.8261} \\
	{\color[HTML]{000000} } &
	{\color[HTML]{000000} Worst} &
	{\color[HTML]{000000} 0.7826} &
	{\color[HTML]{000000} 0.6058} &
	{\color[HTML]{000000} 0.7203} &
	{\color[HTML]{000000} 0.6812} &
	{\color[HTML]{000000} 0.6957} &
	{\color[HTML]{000000} 0.7043} &
	{\color[HTML]{000000} 0.8348} &
	{\color[HTML]{000000} 0.8101} \\
	{\color[HTML]{000000} } &
	{\color[HTML]{000000} Time(s)} &
	{\color[HTML]{000000} 1136.98} &
	{\color[HTML]{000000} 82.23} &
	{\color[HTML]{000000} 815.05} &
	{\color[HTML]{000000} 789.18} &
	{\color[HTML]{000000} 814.23} &
	{\color[HTML]{000000} 798.28} &
	{\color[HTML]{000000} 989.86} &
	{\color[HTML]{000000} 823.65} \\
	{\color[HTML]{000000} } &
	{\color[HTML]{000000} Fitness Rank} &
	{\color[HTML]{000000} 4} &
	{\color[HTML]{000000} 8} &
	{\color[HTML]{000000} 6} &
	{\color[HTML]{000000} 7} &
	{\color[HTML]{000000} 5} &
	{\color[HTML]{000000} 3} &
	\cellcolor[HTML]{E7E6E6}{\color[HTML]{000000} \textbf{1}} &
	{\color[HTML]{000000} 2} \\
	{\color[HTML]{000000} } &
	{\color[HTML]{000000} Time Rank} &
	{\color[HTML]{000000} 8} &
	\cellcolor[HTML]{E7E6E6}{\color[HTML]{000000} \textbf{1}} &
	{\color[HTML]{000000} 5} &
	{\color[HTML]{000000} 2} &
	{\color[HTML]{000000} 4} &
	{\color[HTML]{000000} 3} &
	{\color[HTML]{000000} 7} &
	{\color[HTML]{000000} 6} \\
	{\color[HTML]{000000} HCC Survival} &
	{\color[HTML]{000000} Mean} &
	{\color[HTML]{000000} 0.7196} &
	{\color[HTML]{000000} 0.4974} &
	{\color[HTML]{000000} 0.6489} &
	{\color[HTML]{000000} 0.6574} &
	{\color[HTML]{000000} 0.6985} &
	{\color[HTML]{000000} 0.7105} &
	{\color[HTML]{000000} 0.7317} &
	{\color[HTML]{000000} 0.7105} \\
	{\color[HTML]{000000} } &
	{\color[HTML]{000000} Std} &
	{\color[HTML]{000000} 0.0126} &
	{\color[HTML]{000000} 0.0000} &
	{\color[HTML]{000000} 0.0000} &
	{\color[HTML]{000000} 0.0411} &
	{\color[HTML]{000000} 0.0482} &
	{\color[HTML]{000000} 0.0406} &
	{\color[HTML]{000000} 0.0241} &
	{\color[HTML]{000000} 0.0407} \\
	{\color[HTML]{000000} } &
	{\color[HTML]{000000} Best} &
	{\color[HTML]{000000} 0.7316} &
	{\color[HTML]{000000} 0.4974} &
	{\color[HTML]{000000} 0.6489} &
	{\color[HTML]{000000} 0.7132} &
	{\color[HTML]{000000} 0.7438} &
	{\color[HTML]{000000} 0.7496} &
	{\color[HTML]{000000} 0.7438} &
	{\color[HTML]{000000} 0.7438} \\
	{\color[HTML]{000000} } &
	{\color[HTML]{000000} Worst} &
	{\color[HTML]{000000} 0.6956} &
	{\color[HTML]{000000} 0.4974} &
	{\color[HTML]{000000} 0.6489} &
	{\color[HTML]{000000} 0.5875} &
	{\color[HTML]{000000} 0.6368} &
	{\color[HTML]{000000} 0.6610} &
	{\color[HTML]{000000} 0.6835} &
	{\color[HTML]{000000} 0.6603} \\
	{\color[HTML]{000000} } &
	{\color[HTML]{000000} Time(s)} &
	{\color[HTML]{000000} 1271.23} &
	{\color[HTML]{000000} 114.26} &
	{\color[HTML]{000000} 952.46} &
	{\color[HTML]{000000} 864.63} &
	{\color[HTML]{000000} 872.66} &
	{\color[HTML]{000000} 923.54} &
	{\color[HTML]{000000} 1228.49} &
	{\color[HTML]{000000} 863.92} \\
	{\color[HTML]{000000} } &
	{\color[HTML]{000000} Fitness Rank} &
	{\color[HTML]{000000} 2} &
	{\color[HTML]{000000} 8} &
	{\color[HTML]{000000} 7} &
	{\color[HTML]{000000} 6} &
	{\color[HTML]{000000} 5} &
	{\color[HTML]{000000} 4} &
	\cellcolor[HTML]{E7E6E6}{\color[HTML]{000000} \textbf{1}} &
	{\color[HTML]{000000} 3} \\
	{\color[HTML]{000000} } &
	{\color[HTML]{000000} Time Rank} &
	{\color[HTML]{000000} 8} &
	\cellcolor[HTML]{E7E6E6}{\color[HTML]{000000} \textbf{1}} &
	{\color[HTML]{000000} 6} &
	{\color[HTML]{000000} 3} &
	{\color[HTML]{000000} 4} &
	{\color[HTML]{000000} 5} &
	{\color[HTML]{000000} 7} &
	{\color[HTML]{000000} 2} \\
	{\color[HTML]{000000} Iris} &
	{\color[HTML]{000000} Mean} &
	{\color[HTML]{000000} 0.9720} &
	{\color[HTML]{000000} 0.4267} &
	{\color[HTML]{000000} 0.9733} &
	{\color[HTML]{000000} 0.9667} &
	{\color[HTML]{000000} 0.9740} &
	{\color[HTML]{000000} 0.9733} &
	{\color[HTML]{000000} 0.9773} &
	{\color[HTML]{000000} 0.9733} \\
	{\color[HTML]{000000} } &
	{\color[HTML]{000000} Std} &
	{\color[HTML]{000000} 0.0027} &
	{\color[HTML]{000000} 0.0000} &
	{\color[HTML]{000000} 0.0000} &
	{\color[HTML]{000000} 0.0000} &
	{\color[HTML]{000000} 0.0055} &
	{\color[HTML]{000000} 0.0000} &
	{\color[HTML]{000000} 0.0068} &
	{\color[HTML]{000000} 0.0000} \\
	{\color[HTML]{000000} } &
	{\color[HTML]{000000} Best} &
	{\color[HTML]{000000} 0.9733} &
	{\color[HTML]{000000} 0.4267} &
	{\color[HTML]{000000} 0.9733} &
	{\color[HTML]{000000} 0.9667} &
	{\color[HTML]{000000} 0.9800} &
	{\color[HTML]{000000} 0.9733} &
	{\color[HTML]{000000} 0.9867} &
	{\color[HTML]{000000} 0.9733} \\
	{\color[HTML]{000000} } &
	{\color[HTML]{000000} Worst} &
	{\color[HTML]{000000} 0.9667} &
	{\color[HTML]{000000} 0.4267} &
	{\color[HTML]{000000} 0.9733} &
	{\color[HTML]{000000} 0.9667} &
	{\color[HTML]{000000} 0.9667} &
	{\color[HTML]{000000} 0.9733} &
	{\color[HTML]{000000} 0.9667} &
	{\color[HTML]{000000} 0.9733} \\
	{\color[HTML]{000000} } &
	{\color[HTML]{000000} Time(s)} &
	{\color[HTML]{000000} 912.70} &
	{\color[HTML]{000000} 70.55} &
	{\color[HTML]{000000} 640.51} &
	{\color[HTML]{000000} 628.65} &
	{\color[HTML]{000000} 644.05} &
	{\color[HTML]{000000} 620.96} &
	{\color[HTML]{000000} 898.81} &
	{\color[HTML]{000000} 625.65} \\
	{\color[HTML]{000000} } &
	{\color[HTML]{000000} Fitness Rank} &
	{\color[HTML]{000000} 6} &
	{\color[HTML]{000000} 8} &
	{\color[HTML]{000000} 3} &
	{\color[HTML]{000000} 7} &
	{\color[HTML]{000000} 2} &
	{\color[HTML]{000000} 3} &
	\cellcolor[HTML]{E7E6E6}{\color[HTML]{000000} \textbf{1}} &
	{\color[HTML]{000000} 3} \\
	{\color[HTML]{000000} } &
	{\color[HTML]{000000} Time Rank} &
	{\color[HTML]{000000} 8} &
	\cellcolor[HTML]{E7E6E6}{\color[HTML]{000000} \textbf{1}} &
	{\color[HTML]{000000} 5} &
	{\color[HTML]{000000} 4} &
	{\color[HTML]{000000} 6} &
	{\color[HTML]{000000} 2} &
	{\color[HTML]{000000} 7} &
	{\color[HTML]{000000} 3} \\
	{\color[HTML]{000000} Somerville Happiness} &
	{\color[HTML]{000000} Mean} &
	{\color[HTML]{000000} 0.5688} &
	{\color[HTML]{000000} 0.5114} &
	{\color[HTML]{000000} 0.5386} &
	{\color[HTML]{000000} 0.5114} &
	{\color[HTML]{000000} 0.5555} &
	{\color[HTML]{000000} 0.5386} &
	{\color[HTML]{000000} 0.5615} &
	{\color[HTML]{000000} 0.5386} \\
	{\color[HTML]{000000} Survey} &
	{\color[HTML]{000000} Std} &
	{\color[HTML]{000000} 0.0222} &
	{\color[HTML]{000000} 0.0000} &
	{\color[HTML]{000000} 0.0000} &
	{\color[HTML]{000000} 0.0000} &
	{\color[HTML]{000000} 0.0208} &
	{\color[HTML]{000000} 0.0000} &
	{\color[HTML]{000000} 0.0126} &
	{\color[HTML]{000000} 0.0000} \\
	{\color[HTML]{000000} } &
	{\color[HTML]{000000} Best} &
	{\color[HTML]{000000} 0.5871} &
	{\color[HTML]{000000} 0.5114} &
	{\color[HTML]{000000} 0.5386} &
	{\color[HTML]{000000} 0.5114} &
	{\color[HTML]{000000} 0.5810} &
	{\color[HTML]{000000} 0.5386} &
	{\color[HTML]{000000} 0.5810} &
	{\color[HTML]{000000} 0.5386} \\
	{\color[HTML]{000000} } &
	{\color[HTML]{000000} Worst} &
	{\color[HTML]{000000} 0.5114} &
	{\color[HTML]{000000} 0.5114} &
	{\color[HTML]{000000} 0.5386} &
	{\color[HTML]{000000} 0.5114} &
	{\color[HTML]{000000} 0.5386} &
	{\color[HTML]{000000} 0.5386} &
	{\color[HTML]{000000} 0.5386} &
	{\color[HTML]{000000} 0.5386} \\
	{\color[HTML]{000000} } &
	{\color[HTML]{000000} Time(s)} &
	{\color[HTML]{000000} 851.08} &
	{\color[HTML]{000000} 78.73} &
	{\color[HTML]{000000} 637.46} &
	{\color[HTML]{000000} 643.25} &
	{\color[HTML]{000000} 606.71} &
	{\color[HTML]{000000} 628.12} &
	{\color[HTML]{000000} 895.86} &
	{\color[HTML]{000000} 644.89} \\
	{\color[HTML]{000000} } &
	{\color[HTML]{000000} Fitness Rank} &
	\cellcolor[HTML]{E7E6E6}{\color[HTML]{000000} \textbf{1}} &
	{\color[HTML]{000000} 7} &
	{\color[HTML]{000000} 4} &
	{\color[HTML]{000000} 7} &
	{\color[HTML]{000000} 3} &
	{\color[HTML]{000000} 4} &
	{\color[HTML]{000000} 2} &
	{\color[HTML]{000000} 4} \\
	{\color[HTML]{000000} } &
	{\color[HTML]{000000} Time Rank} &
	{\color[HTML]{000000} 7} &
	\cellcolor[HTML]{E7E6E6}{\color[HTML]{000000} \textbf{1}} &
	{\color[HTML]{000000} 4} &
	{\color[HTML]{000000} 5} &
	{\color[HTML]{000000} 2} &
	{\color[HTML]{000000} 3} &
	{\color[HTML]{000000} 8} &
	{\color[HTML]{000000} 6} \\
	{\color[HTML]{000000} Wine} &
	{\color[HTML]{000000} Mean} &
	{\color[HTML]{000000} 0.9618} &
	{\color[HTML]{000000} 0.7157} &
	{\color[HTML]{000000} 0.6513} &
	{\color[HTML]{000000} 0.9382} &
	{\color[HTML]{000000} 0.9557} &
	{\color[HTML]{000000} 0.9638} &
	{\color[HTML]{000000} 0.9644} &
	{\color[HTML]{000000} 0.9635} \\
	{\color[HTML]{000000} } &
	{\color[HTML]{000000} Std} &
	{\color[HTML]{000000} 0.0055} &
	{\color[HTML]{000000} 0.0000} &
	{\color[HTML]{000000} 0.0000} &
	{\color[HTML]{000000} 0.0116} &
	{\color[HTML]{000000} 0.0075} &
	{\color[HTML]{000000} 0.0050} &
	{\color[HTML]{000000} 0.0024} &
	{\color[HTML]{000000} 0.0027} \\
	{\color[HTML]{000000} } &
	{\color[HTML]{000000} Best} &
	{\color[HTML]{000000} 0.9663} &
	{\color[HTML]{000000} 0.7157} &
	{\color[HTML]{000000} 0.6513} &
	{\color[HTML]{000000} 0.9497} &
	{\color[HTML]{000000} 0.9663} &
	{\color[HTML]{000000} 0.9663} &
	{\color[HTML]{000000} 0.9663} &
	{\color[HTML]{000000} 0.9663} \\
	{\color[HTML]{000000} } &
	{\color[HTML]{000000} Worst} &
	{\color[HTML]{000000} 0.9497} &
	{\color[HTML]{000000} 0.7157} &
	{\color[HTML]{000000} 0.6513} &
	{\color[HTML]{000000} 0.9212} &
	{\color[HTML]{000000} 0.9497} &
	{\color[HTML]{000000} 0.9497} &
	{\color[HTML]{000000} 0.9608} &
	{\color[HTML]{000000} 0.9608} \\
	{\color[HTML]{000000} } &
	{\color[HTML]{000000} Time(s)} &
	{\color[HTML]{000000} 925.41} &
	{\color[HTML]{000000} 108.37} &
	{\color[HTML]{000000} 654.10} &
	{\color[HTML]{000000} 632.19} &
	{\color[HTML]{000000} 649.95} &
	{\color[HTML]{000000} 637.86} &
	{\color[HTML]{000000} 911.85} &
	{\color[HTML]{000000} 626.50} \\
	{\color[HTML]{000000} } &
	{\color[HTML]{000000} Fitness Rank} &
	{\color[HTML]{000000} 4} &
	{\color[HTML]{000000} 7} &
	{\color[HTML]{000000} 8} &
	{\color[HTML]{000000} 6} &
	{\color[HTML]{000000} 5} &
	{\color[HTML]{000000} 2} &
	\cellcolor[HTML]{E7E6E6}{\color[HTML]{000000} \textbf{1}} &
	{\color[HTML]{000000} 3} \\
	{\color[HTML]{000000} } &
	{\color[HTML]{000000} Time Rank} &
	{\color[HTML]{000000} 8} &
	\cellcolor[HTML]{E7E6E6}{\color[HTML]{000000} \textbf{1}} &
	{\color[HTML]{000000} 6} &
	{\color[HTML]{000000} 3} &
	{\color[HTML]{000000} 5} &
	{\color[HTML]{000000} 4} &
	{\color[HTML]{000000} 7} &
	{\color[HTML]{000000} 2} 
	\\ \hline
\end{longtable}
}

\endgroup
\begin{table}[!htb]
	\footnotesize
	\centering
	\caption{Summary of Evaluations for SVMs Optimization}
	\label{tab:svm-summary}
	\begin{tabular}{@{}lcccc@{}}
		\toprule
		{\color[HTML]{000000} Algorithm} &
		{\color[HTML]{000000} \begin{tabular}[c]{@{}c@{}}Avg Fitness \\ Rank\end{tabular}} &
		{\color[HTML]{000000} \begin{tabular}[c]{@{}c@{}}Avg Time \\ Rank\end{tabular}} &
		{\color[HTML]{000000} \begin{tabular}[c]{@{}c@{}}Overall Fitness \\ Rank\end{tabular}} &
		{\color[HTML]{000000} \begin{tabular}[c]{@{}c@{}}Overall Time\\ Rank\end{tabular}} \\ \midrule
		{\color[HTML]{000000} ABC} &
		{\color[HTML]{000000} 3.4} &
		{\color[HTML]{000000} 7.8} &
		{\color[HTML]{000000} 4} &
		{\color[HTML]{000000} 8} \\
		{\color[HTML]{000000} CMA-ES} &
		{\color[HTML]{000000} 7.6} &
		{\color[HTML]{000000} 1.0} &
		{\color[HTML]{000000} 8} &
		\cellcolor[HTML]{E7E6E6}{\color[HTML]{000000} \textbf{1}} \\
		{\color[HTML]{000000} DE} &
		{\color[HTML]{000000} 5.6} &
		{\color[HTML]{000000} 5.2} &
		{\color[HTML]{000000} 6} &
		{\color[HTML]{000000} 6} \\
		{\color[HTML]{000000} GA} &
		{\color[HTML]{000000} 6.6} &
		{\color[HTML]{000000} 3.4} &
		{\color[HTML]{000000} 7} &
		{\color[HTML]{000000} 2} \\
		{\color[HTML]{000000} PSO} &
		{\color[HTML]{000000} 4.0} &
		{\color[HTML]{000000} 4.2} &
		{\color[HTML]{000000} 5} &
		{\color[HTML]{000000} 5} \\
		{\color[HTML]{000000} SSA} &
		{\color[HTML]{000000} 3.2} &
		{\color[HTML]{000000} 3.4} &
		{\color[HTML]{000000} 3} &
		{\color[HTML]{000000} 2} \\
		{\color[HTML]{000000} VSO} &
		{\color[HTML]{000000} 1.2} &
		{\color[HTML]{000000} 7.2} &
		\cellcolor[HTML]{E7E6E6}{\color[HTML]{000000} \textbf{1}} &
		{\color[HTML]{000000} 7} \\
		{\color[HTML]{000000} WOA} &
		{\color[HTML]{000000} 3.0} &
		{\color[HTML]{000000} 3.8} &
		{\color[HTML]{000000} 2} &
		{\color[HTML]{000000} 4} \\ \bottomrule
	\end{tabular}
\end{table}
\section{Conclusion}\label{section:conclusion}
In summary, a novel and powerful meta-heuristic optimization algorithm
called VSO is proposed 
for tackling challenging continuous optimization problems in many real-life applications.
Inspired by the spread and behavior of viruses, the algorithm is carefully devised with
different viral operations to diversify the searching strategies in order
to highly improve its optimizing capacity. 
     
In this paper, VSO is firstly evaluated on a total of $46$ well-known benchmark functions
covering many different types of optimization problems.
The rate of convergence, scalability, and reliability of the algorithm are well-validated
on all these benchmark functions.
Moreover, VSO is used to solve two real-world applications
including the financial portfolio optimization and optimization of hyper-parameters of SVMs for classification problems.
All the obtained results are carefully compared and analyzed with those of classical algorithms
such as GA, PSO, and DE as well as the state-of-the-art optimization approaches including CMA-ES, WOA, and SSA.

The results demonstrate the outstanding performance of our proposed algorithm in terms of
solution fitness, convergence rate, scalability, reliability, and flexibility. Especially, VSO shows a unique potential for high-dimensional continuous optimization problems. Additionally, the algorithmic framework is much flexible to provide an interface to hybridization with other algorithms.
     
The drawbacks of VSO are summarized as follows.
First, the number of algorithmic parameters is larger than those of the existing popular optimization approaches like GA.
Second, the implementation is a bit complicated.
Last but not least,
the computational speed is not mostly in the dominant position due to its  multiple searching strategies.
     
Concerning the future work, how to make the parameters of VSO to be self-adaptive is worth exploring. A more thorough investigation should be conducted on the imported infection operation.
In addition, the applicability of VSO can be further investigated in various real-world applications.
Lastly, VSO has a great potential to be extended for solving mixed continuous-discrete as well as
multi-objective optimization problems.

	
\appendix{}
\section{Classical Benchmarking Functions}\label{appendix:classcial-funcs}
\begin{table}[!htb]
	\footnotesize
	\centerfloat
	\caption{Classical Benchmarking Functions}
	\label{tab:classical_benchmarking_functions}
	\begin{tabular}{@{}lllrc@{}}
		\toprule
		Function & Name                   & Expression & Search Range     & Global Optimum $f(x^*)$                 \\ \midrule
		F1       & Sphere                 &  $ f(\mathbf{x})= {\sum_{i=1}^{D} x_i^{2}}$
		          & {[}-1000,1000{]} & 0                              \\
		F2       & Brown                  &  $f(\mathbf{x})= \sum_{i=1}^{D-1}(x_i^2)^{(x_{i+1}^{2}+1)}+(x_{i+1}^2)^{(x_{i}^{2}+1)}
		$          & {[}-1,4{]}       & 0                              \\
		F3       & Ellipsoid      &        $f(\mathbf{x})=\sum_{i=1}^{D}\left[\left(1000^{\frac{1}{D-1}} x_{i}\right)\right]^{2}$            & {[}-5.12,5.12{]} & 0                              \\
		F4       & Schwefel 2.21          &$f(\mathbf{x})=\max_{i=1,...,D}|x_i| $            & {[}-100,100{]}   & 0                              \\
		F5       & Weighted Sphere        &    $ f(\mathbf{x})= i {\sum_{i=1}^{D} x_i^{2}}$        & {[}-5.12,5.12{]} & 0                              \\
		F6       & Sum of Different Powers & $f(\textbf{x})=\sum_{i=1}^{D}\left|x_{i}\right|^{i+1}$           & {[}-1,1{]}       & 0                              \\
		F7       & Zakharov               &$f(\textbf{x})=\sum_{i=1}^D x_i^{2}+(\sum_{i=1}^D 0.5ix_i)^2 + (\sum_{i=1}^D 0.5ix_i)^4
		$            & {[}-5,10{]}      & 0                              \\
		F8       & Schwefel 1.2           & $f(\textbf{x})=\sum_{i=1}^{D}\left(\sum_{j=1}^{i}x_j\right)^2$         & {[}-100,100{]}   & 0                              \\
		F9       & Rastrigin              & $f(\textbf{x})=10d + \sum_{i=1}^{D}(x_i^2 - 10cos(2\pi x_i))$           & {[}-5.12,5.12{]} & 0                              \\
		F10      & Ackley                 &  $f(\textbf{x})= -20exp(-0.2\sqrt{\frac{1}{D}\sum_{i=1}^{D}x_i^2})$    & {[}-32,32{]}     & 0                              \\
		&&$-exp(\frac{1}{D}\sum_{i=1}^{D}cos(2x_i))+ 20 + exp(1)$      &&\\
		F11      & Griewank               &  $f(\textbf{x}) = 1 + \sum_{i=1}^{D} \frac{x_i^{2}}{4000} - \prod_{i=1}^{D}cos(\frac{x_i}{\sqrt{i}})
		$          & {[}-100,100{]}   & 0                              \\
		F12      & Styblinski-Tank        & $f(\textbf{x}) = \frac{1}{2}\sum_{i=1}^{D} (x_i^4 -16x_i^2+5x_i)
		$           & {[}-5,5{]}       & -39.16599D                  \\
		F13      & Csendes                &  $f(\textbf{x})=\sum_{i=1}^{D}x_i^6(2+sin\frac{1}{x_i})$          & {[}-1,1{]}       & 0                              \\
		F14      & Xin-She Yang N.2&  $f(\mathbf{x})=(\sum_{i=1}^{D}|x_i|)exp(-\sum_{i=1}^{D}sin(x_i^2))
		$          & {[}-2$\pi$,2$\pi${]}   & 0                              \\
		F15      & Alpine N.1              &  $f(\mathbf x)=\sum{i=1}^{D}|x_i sin(x_i)+0.1x_i|
		$          & {[}-10,10{]}     & 0                              \\
		F16      & Michalewicz            & $f(\textbf{x})=-\sum_{i=1}^{D} \sin \left(x_{i}\right) \sin ^{20}\left(\frac{i x_{i}^{2}}{\pi}\right)$           & {[}0,$\pi${]}       & -1.8013 (D=2) \\ 
		&$D\in \{30,100,300,500,1000\}$&&&\\
		\bottomrule
	\end{tabular}
\end{table}
\clearpage
\section{CEC Benchmarking Functions}\label{appendix:cec-funcs}
\begin{table}[!htb]
	\footnotesize
	\centerfloat
	\caption{CEC Benchmark Functions}
	\label{tab:CEC_benchmark_functions}
	\begin{tabular}{@{}llccc@{}}
		\toprule
		Function & Name                                       & Dimension & Search Range   & Global Optimum $f(x^*)$ \\ \midrule
		CEC1     & Rotated High Conditioned Elliptic Function & 30        & {[}-100,100{]} & 100            \\
		CEC2     & Rotated Bent Cigar Function                & 30        & {[}-100,100{]} & 200            \\
		CEC3     & Rotated Discus Function                    & 30        & {[}-100,100{]} & 300            \\
		CEC4     & Shifted and Rotated Rosenbrock’s Function  & 30        & {[}-100,100{]} & 400            \\
		CEC5     & Shifted and Rotated Ackley’s Function      & 30        & {[}-100,100{]} & 500            \\
		CEC6     & Shifted and Rotated Weierstrass Function   & 30        & {[}-100,100{]} & 600            \\
		CEC7     & Shifted and Rotated Griewank’s Function    & 30        & {[}-100,100{]} & 700            \\
		CEC8     & Shifted Rastrigin’s Function               & 30        & {[}-100,100{]} & 800            \\
		CEC9     & Shifted and Rotated Rastrigin’s Function   & 30        & {[}-100,100{]} & 900            \\
		CEC10    & Shifted Schwefel’s Function                & 30        & {[}-100,100{]} & 1000           \\
		CEC11    & Shifted and Rotated Schwefel’s Function    & 30        & {[}-100,100{]} & 1100           \\
		CEC12    & Shifted and Rotated Katsuura Function      & 30        & {[}-100,100{]} & 1200           \\
		CEC13    & Shifted and Rotated HappyCat Function      & 30        & {[}-100,100{]} & 1300           \\
		CEC14    & Shifted and Rotated HGBat Function         & 30        & {[}-100,100{]} & 1400           \\
		CEC15 & Shifted and Rotated Expanded Griewank’s plus Rosenbrock’s Function & 30 & {[}-100,100{]} & 1500 \\
		CEC16 & Shifted and Rotated Expanded Scaffer’s F6 Function                 & 30 & {[}-100,100{]} & 1600 \\
		CEC17    & Hybrid Function 1                          & 30        & {[}-100,100{]} & 1700           \\
		CEC18    & Hybrid Function 2                          & 30        & {[}-100,100{]} & 1800           \\
		CEC19    & Hybrid Function 3                          & 30        & {[}-100,100{]} & 1900           \\
		CEC20    & Hybrid Function 4                          & 30        & {[}-100,100{]} & 2000           \\
		CEC21    & Hybrid Function 5                          & 30        & {[}-100,100{]} & 2100           \\
		CEC22    & Hybrid Function 6                          & 30        & {[}-100,100{]} & 2200           \\
		CEC23    & Composition Function 1                     & 30        & {[}-100,100{]} & 2300           \\
		CEC24    & Composition Function 2                     & 30        & {[}-100,100{]} & 2400           \\
		CEC25    & Composition Function 3                     & 30        & {[}-100,100{]} & 2500           \\
		CEC26    & Composition Function 4                     & 30        & {[}-100,100{]} & 2600           \\
		CEC27    & Composition Function 5                     & 30        & {[}-100,100{]} & 2700           \\
		CEC28    & Composition Function 6                     & 30        & {[}-100,100{]} & 2800           \\
		CEC29    & Composition Function 7                     & 30        & {[}-100,100{]} & 2900           \\
		CEC30    & Composition Function 8                     & 30        & {[}-100,100{]} & 3000           \\ \bottomrule
	\end{tabular}
\end{table}
\clearpage
\section{Full Stocks List}\label{appendix:stocks_list}
\footnotesize{
\begin{longtable}[c]{@{}llllllllll@{}}
	\caption{Stocks List}
	\label{tab:stocks_list}\\
	\toprule
	\multicolumn{10}{c}{S\&P500}                                                            \\* \midrule
	\endfirsthead
	\multicolumn{10}{c}%
	{{\bfseries Table \thetable\ continued from previous page}} \\
	\endhead
	\bottomrule
	\endfoot
	\endlastfoot
	A      & ALB    & ATO    & C      & CMG    & DAL    & ED     & FCX    & GPN    & HST    \\
	AAL    & ALGN   & ATVI   & CAG    & CMI    & DD     & EFX    & FDX    & GPS    & HSY    \\
	AAP    & ALK    & AVB    & CAH    & CMS    & DE     & EIX    & FE     & GRMN   & HUM    \\
	AAPL   & ALL    & AVGO   & CAT    & CNC    & DFS    & EL     & FFIV   & GS     & IBM    \\
	ABBV   & ALLE   & AVY    & CB     & CNP    & DG     & EMN    & FIS    & GWW    & ICE    \\
	ABC    & ALXN   & AWK    & CBOE   & COF    & DGX    & EMR    & FISV   & HAL    & IDXX   \\
	ABMD   & AMAT   & AXP    & CBRE   & COG    & DHI    & EOG    & FITB   & HAS    & IEX    \\
	ABT    & AMCR   & AZO    & CCI    & COO    & DHR    & EQIX   & FLIR   & HBAN   & IFF    \\
	ACN    & AMD    & BA     & CCL    & COP    & DIS    & EQR    & FLS    & HBI    & ILMN   \\
	ADBE   & AME    & BAC    & CDNS   & COST   & DISCA  & ES     & FLT    & HCA    & INCY   \\
	ADI    & AMGN   & BAX    & CDW    & COTY   & DISCK  & ESS    & FMC    & HD     & INFO   \\
	ADM    & AMP    & BBY    & CE     & CPB    & DISH   & ETFC   & FRC    & HES    & INTC   \\
	ADP    & AMT    & BDX    & CERN   & CPRI   & DLR    & ETN    & FRT    & HFC    & INTU   \\
	ADS    & AMZN   & BEN    & CF     & CPRT   & DLTR   & ETR    & FTI    & HIG    & IP     \\
	ADSK   & ANET   & BIIB   & CFG    & CRM    & DOV    & EVRG   & FTNT   & HII    & IPG    \\
	AEE    & ANSS   & BK     & CHD    & CSCO   & DRE    & EW     & GD     & HLT    & IPGP   \\
	AEP    & ANTM   & BKNG   & CHRW   & CSX    & DRI    & EXC    & GE     & HOG    & IQV    \\
	AES    & AON    & BKR    & CHTR   & CTAS   & DTE    & EXPD   & GILD   & HOLX   & IR     \\
	AFL    & AOS    & BLK    & CI     & CTL    & DUK    & EXPE   & GIS    & HON    & IRM    \\
	AGN    & APA    & BLL    & CINF   & CTSH   & DVA    & EXR    & GL     & HP     & ISRG   \\
	AIG    & APD    & BMY    & CL     & CTXS   & DVN    & F      & GLW    & HPE    & IT     \\
	AIV    & APH    & BR     & CLX    & CVS    & DXC    & FANG   & GM     & HPQ    & ITW    \\
	AIZ    & APTV   & BSX    & CMA    & CVX    & EA     & FAST   & GOOG   & HRB    & IVZ    \\
	AJG    & ARE    & BWA    & CMCSA  & CXO    & EBAY   & FB     & GOOGL  & HRL    & J      \\
	AKAM   & ARNC   & BXP    & CME    & D      & ECL    & FBHS   & GPC    & HSIC   & JBHT   \\ \midrule
	\multicolumn{10}{c}{CSI300}                                                             \\ \midrule
	000001 & 000651 & 000983 & 002352 & 300144 & 600066 & 600352 & 600637 & 600900 & 601377 \\
	000002 & 000671 & 002007 & 002385 & 300168 & 600068 & 600362 & 600660 & 600958 & 601390 \\
	000008 & 000686 & 002008 & 002415 & 300251 & 600085 & 600369 & 600663 & 600959 & 601398 \\
	000009 & 000709 & 002024 & 002424 & 300315 & 600089 & 600372 & 600674 & 600999 & 601555 \\
	000060 & 000718 & 002027 & 002456 & 600000 & 600100 & 600373 & 600685 & 601006 & 601600 \\
	000063 & 000725 & 002049 & 002465 & 600008 & 600104 & 600376 & 600688 & 601009 & 601601 \\
	000069 & 000728 & 002065 & 002466 & 600009 & 600109 & 600383 & 600690 & 601018 & 601607 \\
	000100 & 000738 & 002074 & 002470 & 600010 & 600111 & 600406 & 600703 & 601021 & 601608 \\
	000156 & 000750 & 002081 & 002475 & 600015 & 600115 & 600415 & 600704 & 601088 & 601618 \\
	000157 & 000768 & 002131 & 002500 & 600016 & 600118 & 600436 & 600705 & 601099 & 601628 \\
	000166 & 000776 & 002142 & 002508 & 600018 & 600150 & 600446 & 600718 & 601111 & 601633 \\
	000333 & 000783 & 002146 & 002555 & 600019 & 600170 & 600482 & 600737 & 601117 & 601668 \\
	000338 & 000792 & 002152 & 002594 & 600021 & 600177 & 600489 & 600739 & 601118 & 601669 \\
	000402 & 000793 & 002153 & 002673 & 600023 & 600188 & 600498 & 600741 & 601166 & 601688 \\
	000413 & 000826 & 002174 & 002714 & 600028 & 600196 & 600518 & 600795 & 601169 & 601718 \\
	000423 & 000839 & 002183 & 002736 & 600029 & 600208 & 600519 & 600804 & 601186 & 601766 \\
	000425 & 000858 & 002195 & 300017 & 600030 & 600221 & 600522 & 600816 & 601198 & 601788 \\
	000538 & 000876 & 002202 & 300024 & 600031 & 600233 & 600535 & 600820 & 601211 & 601800 \\
	000555 & 000895 & 002230 & 300027 & 600036 & 600256 & 600547 & 600827 & 601216 & 601818 \\
	000559 & 000917 & 002236 & 300033 & 600037 & 600271 & 600549 & 600837 & 601225 & 601857 \\
	000568 & 000938 & 002241 & 300059 & 600038 & 600276 & 600570 & 600871 & 601288 & 601866 \\
	000623 & 000959 & 002292 & 300070 & 600048 & 600297 & 600583 & 600886 & 601318 & 601872 \\
	000625 & 000961 & 002299 & 300072 & 600050 & 600309 & 600585 & 600887 & 601328 & 601877 \\
	000627 & 000963 & 002304 & 300124 & 600060 & 600332 & 600588 & 600893 & 601333 & 601888 \\
	000630 & 000977 & 002310 & 300133 & 600061 & 600340 & 600606 & 600895 & 601336 & 601899 \\* \bottomrule
\end{longtable}
}

\clearpage
\bibliographystyle{unsrt} 
\bibliography{references}
	

\end{document}